\RequirePackage{fix-cm} 
\documentclass{article}
\usepackage{PRIMEarxiv}
\newif\ifsubmission
\ifdefined\submissionmode\submissiontrue\else\submissionfalse\fi

\usepackage[utf8]{inputenc}
\usepackage[T1]{fontenc}
\usepackage{amsmath,amsfonts,bm}
\usepackage{booktabs,multirow,array,longtable,tabularx}
\usepackage[font=normalsize]{caption} 
\usepackage{xcolor}
\usepackage{microtype}
\usepackage{enumitem}
\usepackage{pifont}
\usepackage[authoryear,round]{natbib}
\usepackage{hyperref}
\hypersetup{hidelinks}
\usepackage{lineno}
\usepackage[section]{placeins}

\newcolumntype{L}[1]{>{\raggedright\arraybackslash}p{#1}}
\newcommand{\tableformat}{\fontsize{9.5}{11.5}\selectfont\setlength{\tabcolsep}{3pt}\renewcommand{\arraystretch}{1.15}}
\begin{document}
\raggedbottom

\title{Authorship identification under domain shift: a survey of stylistic measures and learned author representations}

\author{Haining Wang\\
  \normalfont Indiana University School of Medicine\\
  \normalfont Indianapolis, Indiana, USA}

\maketitle
\begingroup
\renewcommand{\thefootnote}{}
\footnotetext{Correspondence: Haining Wang, \href{mailto:hw56@iu.edu}{hw56@iu.edu}.}
\endgroup
\begin{abstract}
Authorship identification uses patterns in writing to infer who wrote a text, but those patterns also reflect topic, genre, and register. This survey argues that topic-independence is not a property of a stylistic feature but of the feature together with its encoding, its scoring rule, and the evaluation split, so the question is which combinations preserve author differences when the domain changes. We organize the evidence in four layers, domains and evaluation conditions, linguistic measures, representations, and scoring and decision rules, and examine within-study comparisons from classical frequency measures to learned author representations and language-model-mediated writing. These comparisons show why feature inventories alone do not explain performance: normalization and encoding change results, author-supervised learning can retain content, and evaluation protocols can reverse model rankings. Successful identification, stylistic similarity, and generalization therefore require separate evidence. The literature offers a practical way to obtain it through designs that separate author identity from particular circumstances of writing: the same authors writing across genres, different authors writing within the same conversation, and parallel texts with controlled content. These designs also help evaluate content masking and language-model rewriting, whose effects depend on what identifying information remains. We use this evidence to guide the choice of stylistic representations and the comparisons needed to establish where they generalize.
\end{abstract}

\section{Introduction}\label{sec:intro}

\emph{Stylometry} is the quantitative study of writing style, using features such as word choice, punctuation, and sentence structure. Its oldest and most developed task is \emph{authorship identification}, which determines who, among a set of candidates, wrote a text of unknown origin. Attribution models with a few hundred hand-crafted features exceed 90\% accuracy with 50 candidates given a few thousand words of reference text per author \citep{abbasi2008writeprints} and place the correct author first among 100,000 blog authors in about one case in five \citep{narayanan2012feasibility}. In the learned-representation era, models distinguish tens of thousands of Amazon reviewers from their product reviews \citep{zhu2021idiosyncratic}. Reviews of the field agree that authorial style is a measurable and discriminative object \citep{holmes1998evolution, juola2008authorship, koppel2009computational, stamatatos2009survey}.

These results depend on the relationship between the reference writings and the questioned text. Differences in topic, genre, platform, period, or editing history can reduce identification performance. Two decades of work on \emph{cross-domain} authorship analysis have documented these performance losses without settling their causes. When a character-trigram classifier drops from 100\% to 44.6\% in a controlled experiment that swaps the training and test domains \citep{bischoff2020importance}, when a leading retrieval model scores poorly on a separate content-independence probe \citep{riverasoto2026steb}, or when a model built from content words alone reaches an area under the receiver operating characteristic curve (AUC, also written AUROC) of 0.95 on a benchmark labeled cross-topic \citep{sawatphol2024addressing}, it is unclear whether the model is recognizing the author or the conditions under which the author wrote.

We argue that no stylistic feature should be assumed to identify authors independently of topic. The question is which combinations of features, representations, and scoring rules preserve author differences when the domain changes. We address it through the authorship-identification pipeline: the linguistic measures extracted from a text, the representation that encodes those measures, and the scoring and decision rule that turns representations into an answer. The literature points to a practical solution in evaluation designs that separate author identity from particular circumstances of writing. Comparing the same authors across genres, different authors within a conversation, and parallel texts with controlled content tests different explanations of an identification result \citep{iarpa2024hiatus, wegmann2022same, liu2026style}. We treat classifiers and neural architectures only where they cannot be separated from the representation they induce, which in the learned-representation literature is often. Defenses against authorship identification, that is, deliberate rewriting to defeat attribution, are outside our scope except where the rewriting is the source of a distribution shift; they are surveyed elsewhere \citep{wang2025dissertation}.

\subsection{Developments in representation and evaluation}

Three developments motivate a closer examination of how stylistic representations generalize across domains. Learned author representations have become the default in new work \citep{riverasoto2021learning, huertastato2022part, patel2025styledistance, patel2023lisa, alkiek2025neurobiber}. What such a vector encodes has become an empirical question, as probing studies find that attribution accuracy and stylistic content are only loosely coupled \citep{terreau2021writing, wang2023authorship, riverasoto2026steb}. Evaluation studies have also examined what these benchmarks measure. The PAN shared tasks on authorship verification, part of the Conference and Labs of the Evaluation Forum (CLEF), defined domain as fan-fiction fandom, which the organizers describe as ``a mixture of topic and literary genre'' \citep{kestemont2020overview}. Leakage audits show that conventional cross-topic splits contain train--test topic similarities above 0.960 \citep{sawatphol2024addressing}. Protocol changes such as how long texts are chunked reverse model rankings on the same data \citep{riverasoto2026steb, kiefer2026when}. Large language models have entered the pipeline at both ends, as editors of human text \citep{uluslu2026impact, malik2026assistant} and as writers with a footprint of their own \citep{shan2026writers, kumarage2023neural}. We treat the survival of human-authorship representations under these interventions as a generalization question of the same kind as topic or genre shift.

\subsection{Organization and contributions}

We organize the evidence in four layers, each a place where generalization can fail, with a reported result depending on all four (Table~\ref{tab:layers}). The \emph{domains and evaluation conditions} layer concerns the circumstances under which a writing sample is produced and observed: topic, genre and register, input mode, time, language, machine mediation, and, on the observation side, text length and candidate-set size (Section~\ref{sec:factors}). The \emph{measure} layer concerns the linguistic quantities extracted from a text, from function-word frequencies to parse-tree paths (Section~\ref{sec:measures}). The \emph{representation} layer concerns how measures are encoded and aggregated, whether as a normalized frequency vector, a per-author language model, or a learned embedding (Section~\ref{sec:representations}). The \emph{scoring and decision rules} layer concerns how representations are compared and used to reach an answer: a classifier, a distance, an impostor ranking, a Bayes factor, an abstention rule, or a prompted language model. These are analytical distinctions, since a method can change more than one layer. Section~\ref{sec:readouts} examines the controlled study in which the scoring mechanism used during training changes what the representation learns. Section~\ref{sec:generalization} assembles within-study contrasts across the four layers and asks what survives a shift, while Section~\ref{sec:open} states the experiments that would separate the explanations left open.

Relative to prior surveys (Table~\ref{tab:surveys}), this review connects feature families and representations to the domain shifts under which they have been evaluated. The family-by-family treatment explains what each measure captures; the cross-domain synthesis asks which of those signals persist. This organization makes it possible to compare conflicting findings under their actual evaluation conditions. It reports evidence as within-study contrasts with the corpus, candidate count, text length, and metric attached, without comparing accuracies across studies. It covers the classical, neural, and language-model eras under one framework, including the negative and null findings that method papers mention in passing. \citet{halvani2019assessing} found that twelve verification methods all performed near chance under topic mismatch. \citet{petropoulos2023contrastive} reported AUC 0.53 on the official PAN 2023 cross-discourse test. \citet{sundararajan2018represents}, often cited for the topic-independence of syntax, reported syntactic features losing on three of four tasks.

\subsection{Scope of the evidence}\label{sec:protocol}

The review covers empirical studies of authorship attribution and verification, together with the linguistic and forensic literature on idiolect and register, in which a representation of authorial style is evaluated under a change of domain or evaluation condition between reference and test, or in which the content of a representation is probed directly. Studies of defenses against authorship identification, of machine-text detection, and of native-language identification are included only where they bear on these questions, and each is labeled by task where it is used. Records were assembled in two rounds, a systematic search with citation chaining closed on 1 September 2026 and a targeted search of six strands on 17 September 2026, with overlapping records across strands. The search logs, reading status of each record, and supporting source passages are kept with the manuscript source. A record enters the text only if its primary page was retrieved and its metadata confirmed. Reported numbers are traced to that page, and calculations made for this review are identified as such. AUC and AUROC values are rounded to two decimal places, with differences smaller than 0.01 described explicitly. A few frequently cited works could be read only in abstract, among them Burrows's Delta paper, the founding stylome study of \citet{vanhalteren2005machine}. Where a number from such a record is used, the text says so. The studies reviewed differ in corpus, candidate count, text length, metric, and the definition of the shift, so no quantitative pooling is attempted. The evidence tables retain each comparison's task and study, distinguish controlled contrasts from cross-task successor comparisons, and do not rank absolute scores across studies.

\begin{table}[t]
\centering\tableformat
\begin{tabular}{L{0.22\linewidth}L{0.7\linewidth}}
\toprule
\textbf{Layer} & \textbf{Where generalization can fail, with one example each} \\
\midrule
Domains and evaluation conditions & The reference writings and the questioned text differ in topic, genre, register, mode, time, language, or editing history, or in how much text and how many candidates are observed. Character trigrams fall from 100\% to 44.6\% under a domain swap \citep{bischoff2020importance}. \\
Measure & The linguistic quantity extracted reflects the domain as well as the author. In a two-author Greek newspaper corpus, four of ten frequent function words carry significant topic effects \citep{mikros2007investigating}. \\
Representation & The encoding determines which relations among measures are retained. Function-word adjacency networks cut error by 23--38\% relative to separately tuned frequency baselines, depending on candidate count; the word inventories were not held fixed \citep{segarra2015authorship}. \\
Scoring and decision rules & The scoring rule reads the representation differently, and in learned systems also shapes what is learned. Models from one backbone trained with mean pooling versus late interaction: recall at 20 (R@20) 0.121 versus 0.497 \citep{kulumba2026where}. \\
\bottomrule
\end{tabular}
\caption{The four layers used to organize this review. R@$k$ denotes recall among the top $k$ retrieved items. A reported accuracy depends on all four; a change at any layer can produce or remove an apparent generalization result.}
\label{tab:layers}
\end{table}
\section{Background: tasks, terms, and domain shift}\label{sec:background}

\subsection{Tasks and decision rules}

Authorship identification is usually posed as a supervised classification problem. Let $X$ be the space of representations extracted from writing samples and $Y$ the label space. Given training data $D=\{(\bm{x}_i, y_i)\}_{i=1}^{n}$, a classifier $f_{\bm{\theta}}$ is fitted by minimizing a loss $\ell$ over $D$,
\begin{equation}\label{eq:classification}
\arg\min_{\bm{\theta}} \sum_{(\bm{x}_i,y_i)\in D} \ell\big(f_{\bm{\theta}}(\bm{x}_i),\, y_i\big).
\end{equation}
Here $n$ is the number of training examples and $\bm{\theta}$ denotes the fitted parameters. In \emph{attribution}, $y_i$ is an author identity drawn from a closed candidate set. In \emph{verification}, each input represents a pair of texts, or a questioned text and a candidate's reference writings, and $y_i$ is binary and encodes whether two texts share an author, or whether a questioned text was written by one named candidate \citep{koppel2012fundamental}. The closed-set assumption is often false for historical, forensic, or web-scale problems, so open-set variants add a rejection option: the model may answer ``none of the candidates'' \citep{narayanan2012feasibility}, abstain when confidence is low \citep{noecker2012distractorless, xie2022many}, or return ``undecidable'' from a separate out-of-distribution detector \citep{boenninghoff2021o2d2}. Verification is the natural formulation when only part of the candidate universe is known.

Equation~\eqref{eq:classification} leaves implicit the scoring and decision rule applied to the representation. In practice the decision is reached in several distinct ways, and the choice matters for generalization. A \emph{discriminative classifier} (logistic regression, SVM, a fine-tuned transformer head) learns a boundary in $X$. A \emph{similarity-based method} compares vectors by a distance or cosine, and is the basis of the Delta family, whose original form averages absolute differences between standardized (z-scored) relative frequencies of frequent words \citep{burrows2002delta, evert2017understanding}, of Siamese and contrastive encoders \citep{boenninghoff2019explainable, riverasoto2021learning}, and of retrieval-style attribution in which the query text is ranked against a pool of candidate documents. The \emph{impostors method} asks not how similar two texts are but whether they are more similar to each other than to a background population \citep{koppel2014determining, seidman2013authorship, potha2017improved}, in which the impostor pool's topical composition becomes part of the decision. A \emph{degradation-based method} measures how quickly a classifier trained to separate two texts collapses as its strongest features are removed. This is unmasking \citep{koppel2004authorship}, which discards the features that carry topic and so reduces topic sensitivity. The original method requires book-length input, and a later generalization brings the requirement down to about four printed pages \citep{bevendorff2019generalizing}. A \emph{generative scoring method} scores a text under a per-author language model and attributes to the lowest cross-entropy, from compression models \citep{teahan2003using} through multi-headed character RNNs \citep{bagnall2015author} to fine-tuned causal language models \citep{huang2024alms}. A \emph{score-calibration method} converts a score to a likelihood ratio or Bayes factor, as in forensic text comparison \citep{ishihara2026chatgpt} and in the probabilistic linear discriminant analysis (PLDA) layer that won the 2020 PAN task \citep{boenninghoff2020deep}. A \emph{prompted language model} compares texts and answers an authorship question in words \citep{hung2023who, huang2024large}. We return to scoring and decision rules in Section~\ref{sec:readouts}. Two systems can therefore share a representation and generalize differently because they score it differently.

A second distinction concerns what is represented. In \emph{instance-based} approaches each document is a training instance; in \emph{profile-based} approaches all of an author's known writings are pooled into one cumulative representation \citep{stamatatos2009survey}. Profile aggregation reduces variance and can substitute for text length. Profile-based extrinsic verifiers are reported to be ``robust when text length is reduced'' \citep{potha2020improved}, and under the cross-fandom open-set conditions of PAN 2019 an ensemble combining both views placed second, 12.6 macro-F1 points above the baseline SVM \citep{bacciu2019cross}. Learned author representations make the same choice explicitly: LUAR combines fixed-length excerpts by attention into one vector for a collection of an author's texts \citep{riverasoto2021learning}, so the representation depends on how the available text is sampled and aggregated, and one candidate with a single long document is not represented the same way as another with many short messages.

\subsection{Premises}

Not every authorship question can be answered by stylometry. Feasibility depends on the writing available. Applying the results below requires examining four premises:
\begin{itemize}
\item The standard setup assumes single-authored questioned and reference texts. Collaborative attribution exists but is understudied, and disentangling authorial components is hard \citep{kestemont2015collaborative, koppel2011unsupervised, dauber2017stylometric}.
\item Adequate reference writing and questioned text are needed to estimate the chosen measures. The amount depends on the measures and the task (Section~\ref{sec:factors-length}).
\item The questioned text should be roughly aligned with the reference writings in topic, genre, register, mode, and time. Differences in these factors can affect the estimated profile.
\item Editing by another person or a language model changes the available evidence, with different effects for correction and extensive rewriting (Section~\ref{sec:factors-mediation}).
\end{itemize}

\subsection{Style, idiolect, and register}\label{sec:terms}

We use \emph{writing style} for the observable ensemble of formal choices in a text: which words, which constructions, which punctuation, in what proportions. We use \emph{authorial style} for the choices a writer makes repeatedly across their texts; how stable those choices are across situations, and how much of them belongs to the situation rather than to the writer, are the empirical questions of Sections~\ref{sec:factors} and \ref{sec:generalization} and are not settled by the definition. A \emph{stylometric profile} or \emph{stylistic representation} is an estimate of authorial style computed from the writings available to an analyst. As a statistical analogy, an author's output over a period is a population and the available writings a sample. The profile estimates a population quantity from a sample, inheriting the sample's composition.

\emph{Idiolect} is the sociolinguistic and forensic term for an individual's distinct version of a language, and its status is disputed. Coulthard writes that ``every native speaker has their own distinct and individual version of the language they speak and write'' and, in the same section, that ``no one has even begun to speculate about how much and what kind of data would be needed to uniquely characterise an idiolect'' \citep{coulthard2004author}. The founding study of the ``human stylome'' hypothesis \citep{vanhalteren2005machine}, read here in abstract, hedges in the same way, placing the number of traits needed for successful distinction ``an order of magnitude larger than assumed so far.'' Reviews of the construct describe it as ``a theory that is not easily observable or measurable, and for which there is little agreement and even less empirical evidence'' \citep{wright2013stylistic} and note that its existence ``may not be empirically provable nor falsifiable'' \citep{grant2020language}. The eliminative position holds that language learning is ``outward bound'' toward the community pattern so that ``individual variation is reduced below the level of linguistic significance'' \citep{labov2012what}. Nini's formalization defines an idiolect not as a fixed grammar but as an indexed family of grammars over time, $I_A=(G^1_A, G^2_A,\ldots,G^T_A)$, where $G^t_A$ is author $A$'s grammar at time index $t$ and $T$ is the number of time indices considered, and states that the fingerprint metaphor ``is only valid if we consider a specific time window'' \citep{nini2023theory}. Temporal drift, which we review in Section~\ref{sec:factors-time}, is therefore predicted by this definition rather than a finding against it.

\emph{Register} and \emph{genre} are the situational variables. Following \citet{biber2019register}, register is variation in language associated with a communicative situation (purpose, participants, medium), and genre is the conventional form a text takes (the novel, the opinion column, the product review). The two overlap in practice and we do not always separate them. They matter theoretically because of an argument due to \citet{grieve2023register}, which accepts that stylometry works and disputes what it measures: variationist sociolinguistics ``cannot provide a theoretical basis for stylometry,'' because analyzing raw frequencies of forms violates its principle of accountability, and what stylometry detects is \emph{individual register variation}, since ``all authors use slightly different registers of the same language in a given communicative context.'' The demonstration is two opinion columnists from similar backgrounds writing for the same newspaper, separated mainly because they chose to write in slightly different registers, with function-word and multidimensional grammatical analyses recovering the same separation. The proposal that Biber's multidimensional analysis could ground authorship analysis is older \citep{nini2013bridging}. The filtering step it implies has a hand-built precedent in corpus linguistics. To isolate genuine ``Blairisms'' from a three-million-word corpus, \citet{mollin2009entirely} had to filter candidate collocations by synonym preference and register specificity, because most collocations that look idiolectal are register artifacts. Content masking and disentangled representations (Sections~\ref{sec:content} and \ref{sec:learned}) are attempts to automate that filter.

We do not adjudicate between the idiolect and register accounts, but we keep them apart because they offer different explanations of the variation attributed to authors. Comparing writers within a register and following the same writers across registers would show which differences persist when the communicative situation changes.

A further terminological difficulty is the word ``style'' in the text style transfer literature, where it names a controllable attribute such as formality, politeness, or sentiment \citep{jin2022deep}. Several style-embedding benchmarks inherit that definition \citep{wegmann2021does, patel2025styledistance, riverasoto2026steb}, although it is not the same object as authorial style. A representation that scores well on formality discrimination has not thereby been shown to identify authors, and vice versa. The Style Text Embedding Benchmark (STEB) reports the two separately, as an ``operational'' and a ``definitional'' score \citep{riverasoto2026steb}.

\subsection{How ``cross-domain'' has been used}\label{sec:shift}

The literature uses ``cross-domain'' and ``domain shift'' for several different changes between reference and test texts, and the differences account for many apparent disagreements. We distinguish these changes from evaluation conditions such as text length and candidate-set size, which can vary without a change of domain.

\emph{Topic.} The subject matter differs between reference and test while genre and register are held constant, as in the Guardian corpus of 13 columnists writing in four thematic sections \citep{stamatatos2013robustness} or the CMCC corpus of 21 students writing on six assigned topics \citep{goldstein2009person}. \emph{Genre and register.} The conventional text form or communicative situation differs: prose against plays \citep{kestemont2012cross}, news against social media \citep{overdorf2016blogs}, or the five genres each written by every foreground author in the HIATUS Resource Sets (HRS), collected for IARPA's Human Interpretable Attribution of Text using Underlying Structure (HIATUS) program \citep{iarpa2024hiatus}. \emph{Mode.} The input mode or environment differs: browser text entry against a word processor \citep{wang2021mode}, mobile against desktop \citep{overdorf2016blogs}, or writing against transcribed speech \citep{aggazzotti2024can}. \emph{Time.} The reference and test writings are separated by years \citep{cafiero2025this, kiefer2026when}. \emph{Language.} The reference and test writings are in different languages, whether through professional translation \citep{bogdanova2014cross}, machine translation \citep{mikros2024cross}, or shared annotation schemes, supplemented in one study by a bilingual pseudolemma glossary \citep{murauer2021dt, cinkova2020stylometry}. \emph{Length and candidate-set size.} The amount of text per decision or the number of candidates changes; these are not nuisance parameters, since the same representation moves from near 100\% to 46\% accuracy when candidates go from two to 10,000 \citep{koppel2013authorship}. \emph{Machine mediation.} A language model has corrected, edited, paraphrased, translated, or wholly generated the text \citep{uluslu2026impact, malik2026assistant, shan2026writers}.

Many benchmarks blend axes. The PAN cross-domain tasks of 2018--2021 operationalize domain as fan-fiction fandom, which the organizers themselves call ``roughly \ldots a mixture of topic and literary genre'' \citep{kestemont2020overview}, and the 2018 overview describes the task as ``cross-topic and cross-(sub-)genre'' \citep{kestemont2018overview}. A high score on such a task is evidence about transfer across fandoms, not about topic-invariance, and Section~\ref{sec:benchmarks} shows that within these tasks same-author pairs are more topically similar than different-author pairs \citep{kestemont2021overview}. Where a study's shift axis is blended we say so.

\citet{alsheddi2025writing} review writing style change detection within multi-authored documents, organizing methods by document representation and prediction, with separate treatment of datasets and evaluation. That task locates authorial changes within a document; our focus is authorship attribution and verification, and how their representations generalize across domains.

\begin{table}[!htbp]
\centering\tableformat
\caption{Prior surveys and position papers, and what they organize the field by. ``Shift'' indicates whether cross-domain generalization is a primary theme. ``Learned'' indicates coverage of learned author or style representations.}
\label{tab:surveys}
\begin{tabular}{@{}L{0.22\linewidth}L{0.23\linewidth}L{0.08\linewidth}L{0.09\linewidth}L{0.32\linewidth}@{}}
\toprule
\textbf{Survey} & \textbf{Organizing axis} & \textbf{Shift} & \textbf{Learned} & \textbf{Note} \\
\midrule
\citet{juola2008authorship} & Feature, then attributional analysis & no & no & Separates feature choice from the decision rule; transfer treated as a forensic special case \\
\citet{stamatatos2009survey} & Feature type; profile vs.\ instance & partial & no & Introduces the profile/instance distinction used here \\
\citet{koppel2009computational} & Problem type (few/many candidates, open set) & partial & no & Difficulty as a property of the task \\
\citet{neal2017surveying} & Application & no & no & Extends stylometry to identity-level tasks \\
\citet{lagutina2019survey} & Feature inventory by linguistic level & no & no & Closest existing representation taxonomy, hand-crafted only \\
\citet{argamon2019forensic} & Assumptions and required controls per method & partial & no & Closest precedent for an assumptions-first framing \\
\citet{grant2022idea} & Forensic validation & partial & no & Validation attaches to a protocol for a case question, not to a system \\
\citet{tyo2022state} & Benchmark reimplementation & partial & yes & Cross-dataset comparison under one protocol \\
\citet{he2024authorship} & Method chronology & partial & yes & General methods survey \\
\citet{alsheddi2025writing} & Representation and prediction for style change detection & partial & yes & Multi-author segmentation; topic and language limitations discussed \\
\citet{huang2026authorship} & LLM-era problems & partial & yes & LLM-generated and LLM-assisted text \\
\citet{wegmann2026survey} & Style representation & partial & yes & Representation-side; defines style operationally \\
This survey & Type of shift, across four layers & yes & yes & Within-study contrasts; classical to LLM era \\
\bottomrule
\end{tabular}
\end{table}
\section{Writing style as identifier}\label{sec:identifier}

A writer has considerable freedom in wording, sentence structure, and rhetoric when conveying a fixed meaning. The following sentences are close paraphrases of one another \citep{hoover1999language}:
\begin{itemize}[nosep]
    \item We were at a loss to find a suitable attendant for her.
    \item We could not find the right care-giver for her.
    \item Finding her a satisfactory attendant had us in a predicament.
\end{itemize}
Among the alternatives a writer prefers some over others. Over time, an active vocabulary, preferred constructions, and habits of arrangement combine into recurring choices. Stylometry rests on the claim that these recurrences are stable enough within a writer and different enough between writers to support identification.

The claim has been demonstrated under controlled conditions and at scale. In a controlled experiment, eight Dutch literature students each wrote nine texts across three genres on fixed topics. Pairwise attribution was evaluated with topic- and genre-matched leave-one-out cross-validation. Accuracy reached 81.5\% with 60 function words and 88.1\% with 50 function words and eight punctuation marks \citep{baayen2002experiment}. In a fully crossed design of 21 undergraduates at a U.S. university writing in six genres on six topics, with every cell filled, cross-topic attribution reached 94\% \citep{goldstein2009person}.

At the other end of the scale, representations learned with content words masked distinguished same-author from different-author pairs in an Amazon test split of 64,248 users with verification F1 of 0.775 (SBERT) and 0.786 (SRoBERTa) \citep{zhu2021idiosyncratic}. A 1,188-feature model placed the correct blogger first among 100,000 in 20\% of cases given three test posts \citep{narayanan2012feasibility}. In corpus linguistics, six White House press secretaries performing the same institutional role in the same register remained ``clearly recognisable,'' with idiolectal preferences ``maintained over a period of at least a year or two'' \citep{barlow2013individual}. When 18 writers each summarized the same source text twice, thirty days apart, every writer produced at least one character 7--9-gram that occurred in both of their own summaries and nowhere else in the data, though the authors conclude that individuality lies in ``the combination of a large number of core grammatical constructions'' rather than in isolated phrases \citep{barlow2025corpus}.

The consistency of writing style across an author's documents is not absolute. The similarity between a new document and the writer's earlier work depends partly on the conditions under which both were produced. \citet{overdorf2016blogs} found that a classifier performed better when the gap between training and test samples, measured in feature space, was smaller. A classifier that requires perfect alignment is of little use. The rest of this survey is about what happens as alignment is relaxed.

The organizers of PAN caution that near-ceiling verification results on fan fiction ``may have given the false impression that authorship verification is an almost solved problem'' \citep{bevendorff2022overview}. A best AUC of 0.99 on unseen authors and unseen fandoms \citep{kestemont2021overview} measures transfer within one platform and one genre. On a 2026 benchmark spanning ten languages and 66 genres, the best of 50 models reaches Success@5 (true author in the top five) of 0.258 \citep{huang2026authbench}. The two figures are not in conflict, because they measure different shifts. Section~\ref{sec:benchmarks} examines what such evaluations vary.
\section{Conditions that shape a writing sample and its analysis}\label{sec:factors}

Authorial style is not observed directly. What an analyst observes is a set of texts each produced under a topic, in a genre and register, through an input device, at a time, in a language, at some length, and with some degree of outside editing. A profile estimated from those texts inherits their composition. If a writer's available texts are about theater, their profile overrepresents the character bigram ``th'' relative to the same writer's output on other subjects, and a classifier using that profile will favor test texts about theater or theory over other texts by the true author. Reducing this bias is the point of most of the work reviewed later. Six of the factors below concern domain differences in how the text was produced (topic, genre and register, mode, time, language, and machine mediation). The remaining subsection concerns evaluation conditions, how much text is observed and how the decision is posed (text length and candidate-set size), which change the identification problem rather than the writing. Table~\ref{tab:factors} collects the within-study contrasts.

\subsection{Topic}\label{sec:factors-topic}

The influence of topic on the linguistic measures used to estimate style extends beyond words that name the subject. \citet{mikros2007investigating} examined a balanced corpus of Modern Greek newspaper articles by two authors writing about culture and politics and found topic effects or author-by-topic interactions in familiar stylistic measures, including function-word frequencies, vocabulary richness, and character frequencies. Four of ten frequent function words carried significant topic effects. Sentence length illustrates the interaction: the difference between the two authors reversed direction between culture and politics although sentence length showed no overall topic effect. Their ablation bears directly on the rest of this survey. Restricting the feature set to 22 topic-neutral measures dropped topic classification from 79.5\% to chance (49\%) while author classification fell only from 96\% to 93\%. Their conclusion is that ``many stylometric variables are actually discriminating topic rather than author.''

The persistence of personal patterns across topics nevertheless depends on the reference collection. With genre held constant, \citet{sapkota2014cross} found that training on one topic and testing on another reduced accuracy relative to within-topic evaluation. For the Chat genre with lexical features, accuracy was 25.71\% within topic against 13.11\% across. Pooling several other topics for training recovered the loss and, in several cells, exceeded the within-topic result. For Chat with character n-grams, accuracy was 57.46\% with multiple cross-topic training against 39.21\% within topic, and for Essay 36.66\% against 30.9\%. These pooled-training cells also use more data. With training volume held at four documents per author, multiple-topic training also significantly outperformed single-topic training in all 24 cross-topic comparisons; for character n-grams on Phone, accuracy rose from 18.84\% to 26.35\% (their Table~6). Topic diversity improves cross-topic attribution even at fixed training volume. With more pooled training data, cross-topic performance can also exceed within-topic performance. The PAN 2018 overview provides related evidence in a different setting, a mild but significant correlation ($r=0.200$, $p=0.001$) between the number of distinct training fandoms available for an author and that author's test F1 \citep{kestemont2018overview}.

The problem carries over to learned representations, where it enters through the construction of training pairs. In a Reddit verification experiment, a model trained without controlling the topic of different-author pairs reached AUC 0.79 when those pairs could come from anywhere in the data and 0.58 when they were drawn from the same conversation \citep{wegmann2022same}. Restricting different-author comparisons to the same conversation removed the usefulness of topic difference as a cue, so a high verification score can partly reflect what the candidates discuss. This property also occurs in generic language-model embeddings, as shown across twelve embedding models in French and English, where centroid dispersion orders as varied topics and varied styles, then varied topics with the same style, then same topic with varied styles, then same topic and same style. Topic thus displaces embeddings more than style does in nine of twelve French models and more clearly in English \citep{icard2025embedding}.

Three of the studies most cited for the topic axis analyze the same corpus. The CMCC corpus of \citet{goldstein2009person} is the data behind the cross-topic study of \citet{sapkota2014cross} and the masking experiments of \citet{stamatatos2018masking}. Agreement among them should not be read as three independent replications.

\subsection{Genre and register}\label{sec:factors-genre}

Documents of a genre carry the genre's background variation, which extends to grammatical words. In a study of weblogs, diary entries favored first-person forms while entries that selected and commented on external content favored third-person forms and nominal specification. This communicative difference explained patterns that might otherwise have been attributed to the writers' gender \citep{herring2006gender}. In an early pilot with common-word frequencies, the first two principal components mainly reflected register, while author differences appeared on the third. One writer's literary criticism patterned away from his own crime fiction, while crime fiction by different writers patterned together \citep{baayen1996outside}. Nini's review summarizes earlier evidence that ``function words' sensitivity to genre outweighs their sensitivity to an authorial signal'' \citep{nini2023theory}.

In the fully crossed CMCC design, holding out a genre reduced accuracy more than holding out a topic: training on five topics and testing on the sixth gave 94\% while training on five genres and testing on the sixth gave 82\% \citep{goldstein2009person}. Both numbers come from 21 undergraduates writing to assigned prompts, a design that isolates the factor but does not resemble a deployment setting. Cross-genre verification with unmasking on five contemporary English authors reached 96\% accuracy (F1 95) within prose, 84\% (F1 62) within theater, and 77\% (F1 56) across the two. The authors report that the method's parameter settings ``often appear to be genre-specific or even author-specific'' \citep{kestemont2012evaluating, kestemont2012cross}. A domain swap that holds the author set fixed and pairs each author with a fan-fiction domain never seen for that author in training dropped a character-trigram Naive Bayes model from 100\% to 44.6\% and an SVM from 98.9\% to 67.4\%. When the swap was relaxed so that training contained many examples of one author--domain pairing and a few of the reverse, an SVM over text-distorted input, with less frequent words replaced by placeholders, lost only 3.6--3.9 points, because a few reverse examples plus domain suppression removed most of the collapse \citep{bischoff2020importance}.

Register also explains separations between authors writing in the same conventional form. Two columnists for the same newspaper were separated by a function-word analysis and by a multidimensional grammatical analysis in closely related ways. One favored narrative rhetoric with past events and recurring participants, the other expository rhetoric with heavier noun modification \citep{grieve2023register}. The familiar word-frequency differences corresponded to interpretable differences in how the writers argued. Register can also invalidate a reference collection altogether. For nine authors who wrote in both classical and vernacular Chinese, function-word and character-based SVM and logistic-regression models performed well within a register and near chance when trained on classical and tested on vernacular Chinese \citep{wang2021cross}.

Formality, often named as a register variable, has no dedicated manipulation study that this review could locate. The best available evidence is a control: \citet{wang2021mode} measured formality and found it near-identical across input modes, providing no evidence that differences in measured formality explain the mode effect. The study did not isolate an effect of formality itself.

\subsection{Mode}\label{sec:factors-mode}

The input condition shapes the estimate. \citet{overdorf2016blogs} found differences between tweets composed on mobile and desktop devices; the average per-author distance between mobile and desktop writing was about 1.4 times the distance within mobile writing. The platform effect was far larger than the device effect, which the authors call ``small but consistent.'' \citet{wang2021mode} compared pre-existing formal writing from 18 participants with new essays the same participants typed into a browser, using a Bayesian hierarchical model. They found that 12 of 15 common stylometric features were credibly different, with simpler vocabulary, shorter sentences, and less sentence-length variation online. Models trained on the offline writing were less accurate on the online essays than in cross-validation within the offline writing. The comparison captures a change in the writing situation as a whole, input environment and task together.

Writing against transcribed speech is the largest mode shift studied, changing both what is produced and how it is recorded. On human-transcribed conversational speech with trials built so that speaker identity cannot be read off topic, models trained on written text achieve ``surprisingly good performance in certain settings'' but ``perform markedly worse as conversational topic is increasingly controlled'' \citep{aggazzotti2024can}. A contrastive Siamese verifier reached AUC 0.80 on its written subset and 0.62 when the combined system was tested on a mixed set that included cross-mode pairs \citep{petropoulos2023contrastive}. The task overview reported 0.888 for essay--email pairs against 0.535 for interview--email pairs in one system, but identified training--test overlap in the essay--email comparison \citep{stamatatos2023overview}. Each of these contrasts changes the test set or the system as well as the mode, so none isolates the mode effect on its own.

Transcription also contributes identifying information. Automatic speech recognition output can match or exceed human transcripts for attribution because recognition errors themselves correlate with speaker identity \citep{aggazzotti2026impact}. On 11,699 telephone conversations, an interpretable feature union lost only 0.03 AUC (0.86 to 0.83) when same-speaker pairs were drawn from different conversations and different-speaker pairs from the same topic. Transcripts without case or punctuation performed comparably on those trials, although the ordering reversed on the hardest ones. The comparison did not isolate either feature \citep{aggazzotti2025stylometric}. In the spoken register, part of the apparent speaker signal therefore comes from transcription convention and from topic. A mode comparison has to account for both.

The mode effect has a direct consequence for evaluating any intervention. If reference texts were composed offline and test texts collected through an online task, part of an apparent drop in attribution accuracy is present before any intervention is applied \citep{wang2021mode}. Unmodified writing collected under the same conditions as the test texts is the relevant control.

\subsection{Time}\label{sec:factors-time}

Two independent studies in different languages, genres, and tasks agree that time separates a profile from a text. For six nineteenth-century French novelists with the feature set held constant and only the publication-date gap varied, the SVM's mean verification probability fell from 0.949 at gaps of a year or less to 0.811 at fifteen years. In the same study's Bootstrapped Distance Impostors analysis, the Pearson correlation between gap and mean probability was $-0.970$. Later works were verified against earlier references about 1.3 points more readily than the reverse \citep{cafiero2025this}. In a German benchmark of forum posts, reviews, and fan fiction, verification F1 fell from 0.900 for pairs twelve months apart to 0.690 at nine to ten years, with significant negative correlations in every genre \citep{kiefer2026when}. Across twelve classical verification methods, verification remained possible across a 15.6-year mean gap between scientific papers (best c@1 0.792, where c@1 is accuracy with non-answers credited at the answering rate; seven methods above 0.700) \citep{halvani2019assessing}. Drift is author-specific: on a diachronic corpus of 22 American authors, a regression on syntactic features predicted publication year within 4.4--5.9 years RMSE for the authors it was fitted to and their close contemporaries but ``dropped significantly'' for others \citep{klaussner2018diachronic}. A 37-million-word French resource of 381 works by eleven prolific authors now exists for such within-author diachronic study \citep{seminck2021corpus}.

As Section~\ref{sec:terms} noted, this is what the time-indexed definition of idiolect predicts \citep{nini2023theory}. For an author whose writings span decades, the relevant question is which period the estimated profile represents and how far it lies from the questioned text.

\subsection{Language}\label{sec:factors-language}

Cross-language attribution, training in one language and testing in another, is a shift of its own kind, since the reference and test writings share few surface forms. The classical solution is machine translation followed by monolingual attribution. For six nineteenth-century authors with 146 English and 129 Spanish documents, monolingual bag-of-words SVMs reached 95\% (English) and 93\% (Spanish); testing on Spanish after machine translation, word n-grams with Naive Bayes reached 88\%, character n-grams 69\%, part-of-speech (POS) n-grams 59\%, and sixteen abstract ``high-level'' features 79\%, while the same high-level features with no translation fell to 28--31\% \citep{bogdanova2014cross}. Character n-grams performed worse than word n-grams in this translated-text comparison. Their resistance to topic shift should therefore not be assumed to extend to language shift. With the Author's Multilevel N-gram Profile (AMNP), accuracy was 0.350 in Greek monolingual cross-validation and 0.010 when the Greek-trained classifier was tested on English machine translations in another study \citep{mikros2024cross}. That translation preserves author signal at all is a finding in its own right. Burrows's Delta over most-frequent words applied to translated literature ``usually fails to identify the translator and identifies the author of the original instead'' \citep{rybicki2012great}. Machine translations cluster with human translations of the same text except for stylistically complex authors, with sentence-length distribution rather than lexis separating machine from human translation \citep{rybicki2025can}. These results show that author signal can remain in translated texts. They complicate the claim that Delta measures individual production habits, since the recovered ``author'' has passed through another writer's hands.

Shared annotation schemes are the alternative to translation. Dependency-tree n-grams (DT-grams) over Universal Dependencies support cross-language attribution without translation \citep{murauer2021dt}. Ten resource-free quantitative features carried author signal across 400 authors in six languages at 96.66\% accuracy \citep{sarwar2018scalable}. The scheme is not automatically language-neutral, however. On 27 parallel Czech--German text pairs, POS tags plus glossary pseudolemmas matched translations of the same work with 95.6\% success, whereas adding morphological features and syntactic dependencies reduced success to 16.7\%. The authors suggest that morphological attribute values remain language-specific (Czech past-tense verbs mark gender, German ones do not) so the supposedly universal layer can reintroduce language identity into the representation \citep{cinkova2020stylometry}. In small-scale attribution of bilingual social-media authors, machine translation ``drastically'' improved almost all approaches, syntactic features plus translation were best, and ``pre-trained language models are outperformed by traditional models'' \citep{murauer2021small}. Attribution difficulty is also a property of the language: under near-identical conditions on a purpose-built corpus, performance on English was higher than on Greek, ``often highly significantly so'' \citep{juola2019comparative}. Frequent-word Delta performs less well in highly inflected Polish and Latin in another comparison \citep{rybicki2011deeper}. Reported accuracies are therefore not comparable across languages.

Multilingual training, transfer to an unseen language, and cross-language identification are distinct uses of a representation. mStyleDistance adapts a style-feature inventory to each language and includes triplets whose anchor and comparison texts differ in language \citep{qiu2025mstyledistance}. \citet{kim2025leveraging} combine probabilistic content masking with single-language batches so that negatives cannot be separated by language. They evaluate transfer to unseen languages and domains, which does not test whether two texts in different languages share an author.

\subsection{Observation and decision conditions: text length and candidate-set size}\label{sec:factors-length}

Even without a domain shift, reducing text length or enlarging the candidate pool can markedly reduce identification accuracy. A resampling study places the minimum sample for reliable attribution at about 2,500 words for Latin prose and 5,000 for English, German, Polish, and Hungarian novels, with results below 3,000 words described as ``disastrous'' and more than 60\% false attributions at 1,000 words \citep{eder2015does}. A German verification benchmark measures the curve directly: F1 from about 0.780 at 25 words to nearly 1.000 at 900, with Spearman correlations of 0.840--0.940 between length and F1 for most systems \citep{kiefer2026gerav}. Truncating to 250 characters dropped one verifier's c@1 from 0.970 to 0.727 and another's from 0.923 to 0.505 \citep{halvani2019assessing}. Micro-text is not only short. Tweets have far lower sentence-length variance than web prose (SD 6.4 against 21.4), so truncating long documents is not a valid proxy for the short-message register \citep{schwartz2013authorship}. On tweets, a baseline combining character and word n-grams with distinctive patterns \citep{schwartz2013authorship} was compared with a character n-gram CNN in the experiments of \citet{shrestha2017convolutional}. With 50 authors, their baseline and CNN reached 50.7\% and 54.2\% accuracy with 50 training tweets per author, and 71.2\% and 76.1\% with 1,000. With 1,000 authors and 200 training tweets each, they reached 30.3\% and 36.5\%. These comparisons vary the amount of reference writing, not the length of individual tweets.

Candidate-set size is the other axis. With character trigrams on one corpus, two-way attribution reaches about 80\% and 145-way about 11\% \citep{luyckx2010effect}. The same classical representation moves from ``very close to 100\%'' for two-author problems to 46\% for 10,000 blog authors \citep{koppel2013authorship}. A transformer authorship encoder falls from 91.25\% at 10 candidates to 72.39\% at 250 \citep{huertastato2022part}. A prompted LLM falls from Micro-F1 86.67 at 10 candidates to 62.50 at 20 \citep{huang2024large}. These thresholds qualify several LLM-mediation studies in the next subsection whose texts average 190--244 words, as their authors note.

\subsection{Machine mediation}\label{sec:factors-mediation}

The evidence distinguishes at least three interventions: minimal correction, fluency rewriting, and paraphrase or generation. They have different effects on the same text.

One graded-intervention design is a native-language identification study rather than an authorship study, but its operators and content checks offer a design for studying authorship. The same 450 learner essays from nine first-language groups were compared in five conditions spanning four intervention levels: original text, minimal correction performed manually or by an LLM, unconstrained LLM grammatical correction, and paraphrase. BERTScore compared the LLM outputs with the human minimal-edit reference. Manual minimal correction cost 3.8 points of classification accuracy (88.9 to 85.1), which the authors describe as a modest decline indicating that overt errors are not the only cues. Unconstrained grammatical correction, which also permits fluency edits, cost 24.0 points (to 64.9), and paraphrase 60.2 (to 28.7) \citep{uluslu2026impact}. Independently and on a different task, heavy generative rewriting reduced LUAR attribution accuracy on blogs by 66.5 percentage points and a non-generative grammar corrector by 15; light grammar-only prompting reduced a surface-feature SVM by 5.0 points \citep{malik2026assistant}. AI editing and AI generation have opposite-signed effects on text entropy, so ``LLM involvement'' is not one factor \citep{shan2026writers}.

One result in this cluster is a null. On Reuters C50, where each journalist writes a fixed beat, attribution remained stable after heavy rewriting (+1.0 percentage point, not significant), which the authors attribute to topic carrying the identity on that corpus \citep{malik2026assistant}. On such a corpus a rewriting that alters style can leave attribution accuracy unchanged, so accuracy alone does not show how much stylistic signal remains.

Guided rewriting helps identify interventions worth testing. When a small LLM described a text along four dimensions (sentence length, vocabulary, tone, punctuation) and rewrote it to suppress them, attribution F1 fell from 66.45 to 26.02 for a DeBERTa classifier on ten blog authors, against 36.61 for unguided paraphrase, with the same pattern on nine review authors (Table~\ref{tab:factors}). On reviews, a prompt naming only sentence length produced a similar reduction in F1 to that obtained with the full profile, although the study did not isolate sentence-length changes from the other edits produced \citep{khan2026iamnoone}. The rewriting model also matters. A character 3--5-gram SVM at 0.966 accuracy on held-out French literary texts fell to 0.664 on LLM imitations of the same authors. The surviving signal depended on the imitating model (Proust imitations: Mistral 0.844 against GPT 0.490) \citep{icard2026measuring}. Human intervention can move the signal in the other direction. Post-editing an LLM draft increased its LUAR similarity to the writer's unassisted work, while the writers' perceived personalization correlated only weakly ($r=0.244$) with that similarity \citep{baumler2026postediting}.

At the corpus level, a 2004--2025 benchmark finds ``no evidence of a measurable AI-era distribution shift'' in its data \citep{kiefer2026when}. The significant break in the detectability of native language in edited English text is at the neural-machine-translation boundary, not the LLM boundary, with Chinese moving in the opposite direction (F1 0.812 to 0.885) \citep{utami2026still}. An effect measured on individual texts and an effect measured over a corpus are different quantities.

\subsection{Overlapping conditions}

The axes above rarely occur alone. The PAN 2018 cross-domain task was cross-topic and cross-sub-genre at once \citep{kestemont2018overview}. A change of platform is a change of register, mode, and often length together \citep{overdorf2016blogs}. Speech transcription entangles mode with topic and transcriber convention \citep{aggazzotti2024can}. These comparisons make genre and register important controls, not background labels. In the crossed CMCC design, holding out genre cost more than holding out topic (Section~\ref{sec:factors-genre}). A platform change can combine that difference with changes in mode and length. Studies should therefore distinguish a within-author change of writing situation from a change of author population, and identify which factors move together. Section~\ref{sec:generalization} returns to the question of what a benchmark's labels actually vary. Before that, Sections~\ref{sec:measures} and \ref{sec:representations} review the measures and representations whose sensitivity to these factors is at issue.

\begingroup
\tableformat
\begin{longtable}{L{0.11\linewidth}L{0.27\linewidth}L{0.14\linewidth}L{0.26\linewidth}L{0.14\linewidth}}
\caption{Evidence for domains and evaluation conditions. Most rows give within-study contrasts; the Reddit topic-mismatch row reports one condition, and the heavy-rewrite row compares effects across two corpora within one study. These are not isolated causal effects or a cross-study ranking. The Kestemont row comes from the DH2012 companion abstract. ``pp'' is percentage points.}\label{tab:factors}\\
\toprule
\textbf{Factor} & \textbf{Contrast (held constant)} & \textbf{Metric} & \textbf{Baseline $\to$ shifted} & \textbf{Source} \\
\midrule
\endfirsthead
\caption[]{(continued)}\\
\toprule
\textbf{Factor} & \textbf{Contrast (held constant)} & \textbf{Metric} & \textbf{Baseline $\to$ shifted} & \textbf{Source} \\
\midrule
\endhead
Topic & Within-topic vs.\ single cross-topic, Chat, lexical features & accuracy & 25.71 $\to$ 13.11 & \citet{sapkota2014cross} \\
Topic & Single vs.\ multiple cross-topic training, Chat, char n-grams; more training data & accuracy & 27.56 $\to$ 57.46 & \citet{sapkota2014cross} \\
Topic & Different-author pairs from anywhere vs.\ same conversation & AUC & 0.79 $\to$ 0.58 & \citet{wegmann2022same} \\
Topic & Reddit topic mismatch, 12 verification methods & c@1 & best 0.533 & \citet{halvani2019assessing} \\
Genre vs.\ topic & Held-out topic vs.\ held-out genre, 21 authors & accuracy & 94 $\to$ 82 & \citet{goldstein2009person} \\
Genre & Prose within vs.\ across prose/theater, unmasking & accuracy & 96 $\to$ 77 & \citet{kestemont2012evaluating} \\
Domain & Zero-knowledge swap, NB / SVM; high-imbalance swap with distortion, SVM & accuracy & 100$\to$44.6 / 98.9$\to$67.4; $-$3.6 & \citet{bischoff2020importance} \\
Register & Classical vs.\ vernacular Chinese, subsets of a 9-author corpus & accuracy & within-register $\to$ near chance & \citet{wang2021cross} \\
Mode & Essay--Email vs.\ Interview--Email, PAN 2023; first pairing overlaps training & overall score & 0.888 $\to$ 0.535 & \citet{stamatatos2023overview} \\
Mode & Written-subset verifier vs.\ combined system on mixed test set & AUC & 0.80 $\to$ 0.62 & \citet{petropoulos2023contrastive} \\
Mode & Telephone speech, base vs.\ harder positive and negative pairs & AUC & 0.86 $\to$ 0.83 & \citet{aggazzotti2025stylometric} \\
Time & SVM, gap $\le$1 year vs.\ 15 years, 6 novelists & mean probability & 0.949 $\to$ 0.811 & \citet{cafiero2025this} \\
Time & 0--12 months vs.\ 9--10 years, reviews & F1 & 0.900 $\to$ 0.690 & \citet{kiefer2026when} \\
Language & AMNP: Greek monolingual CV vs.\ Greek-trained, English-MT-tested & accuracy & 0.350 $\to$ 0.010 & \citet{mikros2024cross} \\
Language & Monolingual BOW SVM vs.\ MT + char n-gram NB, 6 authors & accuracy & 95 $\to$ 69 & \citet{bogdanova2014cross} \\
Language & POS + pseudolemmas vs.\ + morphology + dependencies, 27 parallel pairs & translation-matching success & 95.6 $\to$ 16.7 & \citet{cinkova2020stylometry} \\
Length & Full vs.\ 250 characters, two verifiers & c@1 & 0.970 $\to$ 0.727; 0.923 $\to$ 0.505 & \citet{halvani2019assessing} \\
Candidates & 10 vs.\ 250 candidates, transformer encoder & accuracy & 91.25 $\to$ 72.39 & \citet{huertastato2022part} \\
Candidates & 2 vs.\ 145 candidates, char trigrams & accuracy & $\approx$80 $\to$ $\approx$11 & \citet{luyckx2010effect} \\
Mediation & Original vs.\ manual minimal / LLM grammatical correction / paraphrase edit (native-language identification) & accuracy & 88.9 $\to$ 85.1 / 64.9 / 28.7 & \citet{uluslu2026impact} \\
Mediation & LUAR, heavy rewrite on blogs vs.\ on C50 & change in attribution accuracy & $-$66.5 pp vs.\ +1.0 pp (not significant) & \citet{malik2026assistant} \\
Mediation & Original vs.\ generic vs.\ style-profile-guided paraphrase, blogs (DeBERTa) / reviews (BERT) & F1 & 66.45 $\to$ 36.61 / 26.02; 76.78 $\to$ 29.37 / 20.76 & \citet{khan2026iamnoone} \\
\bottomrule
\end{longtable}
\endgroup
\section{Stylistic measures}\label{sec:measures}

Researchers have proposed a great many linguistic measures of writing style \citep{rudman1997state}. Many have produced conflicting findings or rest on assumptions that do not hold. An ideal measure would be consistent within an author's work, discriminating between authors, and stable across the factors of Section~\ref{sec:factors} even when estimated from a partial sample. No measure family meets all three conditions. This section reviews the main families and the evidence for each, with attention to the shift condition under which each was tested. Table~\ref{tab:measures} summarizes the evidence with the condition under which each result holds.

\paragraph{Definitions} We use \emph{stylistic measure} for a quantified property of writing, such as word frequency or sentence length. \emph{Feature} is the broader modeling term because it includes categorical indicators and learned embedding coordinates as well as quantities supplied by measures. A \emph{character} is the smallest unit of a text, including letters, digits, punctuation, whitespace, and special markers such as emoji. A \emph{word} is a sequence of characters delimited by whitespace or punctuation. A \emph{token} is a unit defined by a tokenizer, whether a whole word or a subword produced by byte-pair encoding or a unigram language model \citep{sennrich2016neural, kudo2018subword}. An \emph{n-gram} is a sequence of $n$ consecutive units at the character, token, or word level. The families below overlap because a character n-gram may capture punctuation, a function word, or an idiosyncratic spelling. Synonym choice concerns the use of words rather than a separate linguistic level.

\subsection{Shared properties, and one that does not hold}

The statistical convenience of the three most widely used families, function words, common words, and character n-grams, comes from two shared properties \citep{kestemont2014function}. Their frequencies can be estimated reliably because they occur often, and their wide dispersion means that they occur in nearly every document in a language. A third property often attributed to them, independence from content, does not hold for any of the three.

For function words, the Greek study above found that four of ten frequent function words carried significant topic effects, and only a curated 22-feature subset made topic classification fall to chance \citep{mikros2007investigating}. For character n-grams, \citet{sundararajan2018represents} state that they ``represent character, lexical and syntactic information to varying extents'' and ``amplify the presence of possibly unique word/phrase choices''. \citet{stamatatos2009survey} says low-level features ``unavoidably capture thematic information.'' Content words, often excluded from stylistic feature sets, alone reach verification F1 0.795 against 0.786 for the content-masked model on the same Amazon data \citep{zhu2021idiosyncratic}. Across eight novelists content-word-only and function-word-only models are statistically indistinguishable \citep{sari2018topic}. These families are less sensitive to topic than content words, but their residual sensitivity is measurable and sometimes large.

\subsection{Function words}\label{sec:fw}

Function words mark grammatical relations, for example by linking clauses, introducing noun phrases, or expressing tense and modality. This class includes pronouns, conjunctions, prepositions, determiners, auxiliaries, qualifiers, and interrogatives. Their frequencies have been confirmed as style markers in many studies. However, it is inadvisable to trust their distribution blindly \citep{kestemont2014function}. For instance, some function-word patterns attributed to gender reflect genre conventions \citep{herring2006gender}, while their frequencies also vary with topic \citep{biber1998corpus}. Such influences can be reduced by curation, such as removing personal pronouns \citep{burrows1987computation}.

In nineteenth-century prose spanning genres defined by a library catalog, \citet{menon2011domain} found that a stop-word list approximating function words outperformed POS trigrams and common-word trigrams when training and test domains differed. However, across blogs, tweets, and Reddit comments, function words ranked among the top features together with character and POS n-grams and were still ``far from sufficient'' for cross-domain attribution \citep{overdorf2016blogs}. Disambiguating homographic function words (``to'' as infinitive marker against preposition) and attending to the blend of first- and third-person narration had effects that differed across the prose corpora studied \citep{hoover2001statistical}.

Function-word frequencies and adjacency networks use the same class of words but retain different information because the networks encode transitions between words, not just their counts. Function-word adjacency networks, with adaptively selected word inventories and separately tuned frequency-based baselines, reduced attribution error by 38\% for binary attribution and 29\% on average (10.8\% to 8.3\%, a 23\% reduction, in the ten-author case reported in Table~\ref{tab:generalization}). Combining the two encodings approximately halved the error relative to frequency-based methods \citep{segarra2015authorship}. An audit of the method on the early modern play canon found that with the method's own decision criterion enforced instead of ``lowest wins,'' the success rate fell from a claimed 93.6\% to 72.3\%. Per-author validation collapsed for small canons (Chapman 30.8\%, Jonson 43.8\%). A claimed 23-fold disparity in unique function-word pairs between Shakespeare and Marlowe vanished with size-matched corpora \citep{barber2020function}. Barber's audit identified a relaxed decision rule and argued that unequal canon sizes explained the pair-count disparity; it did not replicate the full attribution method.

\subsection{Common words and their n-grams}

A common word is a word that exceeds some frequency or dispersion threshold. The most frequent words are among the longest-used features in the field \citep{burrows1987word, koppel2007measuring, stamatatos2006authorship}. The cutoff matters. In a single-corpus evaluation of forty columnists, a word profile retaining words that appear in at least ten texts per author reached 90\% accuracy for two candidates and 45\% for forty; the best-performing lists ``contain most of the function words, but most of the content words have been stripped away'' \citep{grieve2007quantitative}. Lowering the threshold admitted content words and degraded attribution; raising it left too few words to discriminate. In the Delta family, most distance measures work best with 1,000--2,000 most frequent words, and the number and frequency range of words affected attribution differently across languages and genres \citep{evert2015explaining, rybicki2011deeper}.

Common words have one advantage over a curated function-word list by capturing \emph{ad hoc} functional markers the list misses, such as ``cuz'' for ``because.'' As \citet{hoover2001statistical} observed, very frequent non-function words can be included ``under the assumption that their usage may also be unconscious.'' Extracted without external resources, they suit fast prototyping and low-resource languages. Word n-grams add local order; ``ask what you can do for your country'' and ``ask what your country can do for you'' are identical as bags of words. Common word n-grams are generally kept short in stylometry \citep{coyotl2006authorship, sanderson2006short}, since longer sequences capture more specific phrases and are more sparse. Testing word orders 1--4, \citet{peng2004augmenting} found no gain beyond unigrams under the best-performing smoothing methods in their Greek authorship task and identified sparse-data overfitting as a reason to limit context length. A ``flexible pattern'' family that captures the context in which function words occur was designed for micro-text \citep{schwartz2013authorship}.

Within this family the clearest demonstration that representation rather than measure carries an improvement is normalization. Dividing a word's count by the summed occurrences of the word and its semantic neighbors, obtained from a locally trained GloVe model, instead of by the total number of tokens, raised Cosine Delta F1 from 0.908 to 0.959 on a corpus of novels with the word list held constant \citep{eder2022boosting}; the neighbors need not be synonyms.

\subsection{Character n-grams}\label{sec:charngrams}

Character n-grams have been effective style discriminators since the earliest computational studies \citep{forsyth1996feature, peng2003language}. A method comparing common character n-gram profiles took the highest score on the fixed-topic English essay task and tied for highest on the free-topic task of the 2004 Ad-hoc Authorship Attribution Competition \citep{kevselj2003n, juola2008authorship}. Character bigrams and trigrams were the most successful single family in a 39-measure comparison on one corpus \citep{grieve2007quantitative}.

Why they work is less clear than for function words. They capture ``a bit of everything'' \citep{kestemont2014function}, including common-word distributions, spelling variants (``organization'' against ``organisation''), tense preferences (``-ing'' against ``-ed''), word stems that carry topic (``bio''), and \emph{ad hoc} markers such as emoticons. Their differentiating power has been attributed to their higher frequency relative to word-based measures \citep{kestemont2014function}. Short n-grams are little affected by isolated typing errors \citep{stamatatos2009survey}. They also need no tokenizer, which suits languages without explicit word boundaries. A related cross-domain result comes from a model without a hand-selected feature inventory. A prediction-by-partial-matching (PPM) compression model with a compression-based cosine, without a separate classifier-training step or hand-selected features, placed third of eleven in the PAN 2018 cross-domain task at macro-F1 0.629 against 0.584 for the character n-gram SVM baseline. On the provided training problems used for additional evaluation, its scores ranged from 0.420 for a 20-candidate Polish problem to 0.896 for a five-candidate French problem, varying both language and candidate count \citep{halvani2018crossdomain}. The robustness usually credited to character n-grams is at least partly a property of character-level statistical modeling rather than of a hand-selected inventory.

\citet{sapkota2015not} report the most detailed decomposition of character n-grams that we located. Trigrams were divided into ten categories by position (Table~\ref{tab:charngrams}): affix n-grams (prefix, suffix, space-prefix, space-suffix), word n-grams (whole-word, mid-word, multi-word), and punctuation n-grams (begin, middle, end). In these experiments, word-spanning categories could be omitted without reducing accuracy. Discarding every word category matched or beat using all categories with about a third fewer features (Guardian1 57.0 against 55.6; Guardian2 48.0 against 45.9), and many of the 1,024 possible combinations could not beat affix plus punctuation. The best-performing category differed between conditions: on single-domain CCAT-50, mid-word n-grams were among the best (61.0) and mid-punctuation the worst (12.4). On cross-domain Guardian1, mid-punctuation was best (46.1) and multi-word among the worst (32.2). The ranking of n-gram types therefore depends on the corpus and the shift condition. A follow-up study examined preprocessing and feature pruning. On the Guardian cross-topic corpus, replacing digits and named entities and splitting punctuation raised a multinomial Naive Bayes (MNB) over affix-plus-punctuation trigrams from 58.4\% to 64.3\%; raising the minimum-frequency threshold from 5 to 200 (341 features, about 88\% fewer) raised it to 73.2\%, whereas on single-topic CCAT-10 digit replacement alone gained at most 0.8 percentage points and the full preprocessing bundle reduced accuracy \citep{markov2017improving}. Additionally replacing each of 150 highly frequent words with its own distinct symbol reduced the MNB score from 64.3 to 62.2. Cross-topic generalization here improved as the representation was made smaller.

Character n-grams are not immune to topic. In same-topic attribution, performance improves as more content words are coded as n-grams and is best when all words are \citep{stamatatos2013robustness, stamatatos2017authorship}. When training on politics and testing on society articles, SVM performance peaked at about 3,500 n-grams, with the optimum differing by topic and genre \citep{stamatatos2013robustness}. That paper is the one most cited for the ordering same-topic $>$ cross-topic $>$ cross-genre, but prints no numeric accuracy. Because all its values exist as plotted curves, we cite its ordering rather than its values. Some n-grams are ``associated with particular content words and roots'' \citep{koppel2009computational}. A topic-diverse corpus, a cutoff tuned on a topic-distinct validation set, attention to stem-like n-grams, and small $n$ mitigate the problem. The right $n$ depends on language and corpus \citep{stamatatos2006ensemble}, while inter-word n-grams that cross whitespace capture sequences within and across word boundaries.

\begin{table}[t]
\centering\tableformat
\caption{Categories of character n-grams in \citet{sapkota2015not}. A whitespace is shown as ``\_''.}
\label{tab:charngrams}
\begin{tabular}{llL{0.45\linewidth}l}
\toprule
\textbf{Class} & \textbf{Category} & \textbf{Annotation} & \textbf{Trigram examples} \\
\midrule
\multirow{4}{*}{Affix} & prefix & first $n$ characters of a word & sum, cat, cha \\
 & suffix & last $n$ characters of a word & ary, ies, ter \\
 & space-prefix & begins with a space & \_ca, \_ch, \_us \\
 & space-suffix & ends with a space & ry\_, es\_, er\_ \\
\midrule
\multirow{3}{*}{Word} & whole-word & covers a whole word & the, for \\
 & mid-word & interior of a word, neither first nor last character & umm, mma, mar \\
 & multi-word & spans a word boundary & y\_o, f\_t, e\_c \\
\midrule
\multirow{3}{*}{Punct.} & begin & first character is punctuation & .\_t, ,\_w \\
 & middle & punctuation neither first nor last & y,\_ \\
 & end & last character is punctuation & e\_(, 5). \\
\bottomrule
\end{tabular}
\end{table}

\subsection{Punctuation}

Punctuation is usually folded into character n-grams and deserves separate mention for its role and its transfer behavior. In the ten-text word profile described above, adding punctuation raised forty-candidate accuracy from 45\% to 61\% \citep{grieve2007quantitative}. Among the ten typed categories, punctuation n-grams generalized best across topics, particularly those beginning with or containing a punctuation mark \citep{sapkota2015not}. In the 100,000-blog study, the apostrophe, period, and comma were among the ten most informative features \citep{narayanan2012feasibility}. Punctuation sequences alone, with all words removed, separate authors and genres \citep{darmon2021pull}, although the authors of that 651-author study note that they do not separate author from editor influence. In telephone speech transcripts, normalized and cased, punctuated versions performed comparably on base and hard trials, but the ordering reversed on harder trials. This comparison changes more than punctuation \citep{aggazzotti2025stylometric}. No dedicated study of punctuation-feature transfer across genre or register could be located.

\subsection{Syntax}\label{sec:syntax}

Syntactic analysis assigns words and phrases their part of speech and grammatical function under a set of parsing rules, either as a constituency hierarchy \citep{chomsky1957syntactic} or as head--dependent relations. Shallow parsing chunks phrases and clauses, whereas deep parsing yields a full tree. Syntactic annotation abstracts away some lexical choice without establishing independence from semantics. Rhetoric offers an example of syntactic style at the sentence level. A \emph{loose} sentence states the main clause first and appends detail, whereas a \emph{periodic} sentence delays the main clause (``She felt rejuvenated, walking along the beach \ldots'' against ``Walking along the beach \ldots, she felt rejuvenated'').

Syntactic annotations can be used to represent choices beyond vocabulary. For instance, the use of phrase-structure rules separated two crime novelists \citep{baayen1996outside}, while a combination of token-, phrase-, and analysis-level markers outperformed a frequent-word baseline in Modern Greek \citep{stamatatos2001computer}. However, promising POS results in a controlled writing study \citep{glover1996detecting} did not translate into an advantage over function words in a restricted Flemish corpus \citep{luyckx2005shallow}. Abstracting away words does not guarantee a better identifier.

Tree topology \citep{feng2012characterizing} and local paths such as pq-grams \citep{augsten2008pq, tschuggnall2014enhancing} retain more structure than POS frequencies, although the pq-gram authorship study did not isolate their contribution against a baseline. Learned representations can incorporate that structure. Syntax-tree paths improved a character CNN \citep{zhang2018syntax}, while POS-tag embeddings improved a GloVe-based hierarchical BiLSTM \citep{jafariakinabad2019style}. Structural word embeddings can also be trained on sentence--parse pairs without requiring a parser at inference, although the cited study trained only on novels and did not test cross-domain transfer \citep{jafariakinabad2022self}. The more important question for this review is whether these representations preserve author differences when the domain changes.

The evidence for syntax under shift is mixed. In the positive study, \citet{raghavan2010authorship} found that on four news datasets built on shared topics, a lexical MaxEnt baseline and per-author PCFGs were near parity, while on the one topic-heterogeneous dataset, Poetry, MaxEnt collapsed to 56.36\% and the PCFG family reached 78.18--87.27\%. In these comparisons syntax added little when topic was controlled and a great deal when it was not. In the negative study, \citet{sundararajan2018represents} found that on Guardian10, a PCFG language model reached 24.14\% across topics and 33.33\% across genres against 70.41\% and 69.84\% for a PPM character language model, and in the four-task design usually cited for syntax's topic-independence, syntax lost on three tasks and won on one. Dependency features labeled with POS captured stylistic similarity across different novels by the same author better than the twenty most common words at larger sample sizes, an advantage that may owe something to the small size of the word set \citep{bjorklund2017syntactic}. The most developed recent line treats grammar as a behavioral biometric. A cognitively motivated grammar model with POS-noise masking, which replaces selected content tokens with part-of-speech labels, placed first on 7 of 12 corpora and improved under masking in all 12 (IMDb 0.751 to 0.831) \citep{nini2026grammar}. These methods depend on the parser, since ``a percentage of sentences which the parser fails to analyse'' limits informal text \citep{bjorklund2017syntactic}, and can be sensitive to language differences, as illustrated in Section~\ref{sec:factors-language}, where adding morphological and dependency layers of a supposedly universal scheme reduced translation-matching success from 95.6\% to 16.7\% \citep{cinkova2020stylometry}. No syntactic n-gram study with an explicit cross-topic split could be located \citep{sidorov2013syntactic, posadasduran2025learning}. \citet{glover1996detecting}, drawing on \citet{milic1967quantitative}, proposed that syntactic preferences vary little across domains because writers choose words deliberately and parts of speech unconsciously; the evidence above supports that claim in some settings and not in others.

\subsection{Idiosyncrasies}

Idiosyncrasies are non-standard spelling, formatting, and grammatical choices that recur in a writer's output. They include misspellings, repeated whitespace, unusual punctuation (``!!!!''), all-capital formatting, neologisms (``folx,'' ``smol''), acronym variants, orthographic preferences (``e-mail'' against ``email''), letter--number substitutions (``4get''), accent stylizations (``cuz'') \citep{grant2012txt}, emoticons and emoji, and grammatical quirks such as omitted words, tense mismatches, run-ons, and fragments \citep{koppel2003exploiting}. Some are deliberate choices, such as the serial comma or an aversion to semicolons. Others deviate from acceptability only relative to a community, as American spellings do in British prose. Even scrupulously correct spelling can be a quirk. The perfect spelling of the Unabomber's manifesto pointed to his education \citep{foster2000author}. Forensic linguists have long used such clues \citep{love2002attributing, grant2012txt}.

In the only located ablation, error features with C4.5 combined with either lexical or POS features beat lexical and POS combined without them (68.8 and 71.8 against 61.7) \citep{koppel2003exploiting}. The stated failure mode is sparsity, since ``authors might make it through an entire short document without committing any of their habitual errors.'' The mediation curve of Section~\ref{sec:factors-mediation} bears on this indirectly. In native-language identification, removing overt errors by minimal correction cost 3.8 points while fluency rewriting cost far more \citep{uluslu2026impact}. Whether the same holds for individual authors has not been tested. Orthographic and paralinguistic features can transfer where classical features do not. Twenty emoji-based features on 4,697 Reddit authors reached 0.787 pairwise verification accuracy against 0.873 for a 27-feature classical set and 0.895 combined. Under cross-subreddit train--test the emoji features held ``virtually identical'' accuracy while the classical features fell 26.6 percentage points \citep{etli2026emoji}. The study is single-platform and restricted to emoji-using authors.

\subsection{Synonym choice}

A writer's preferences among near-synonyms can help identify the writer. \citet{koppel2006feature} argued that frequent features with readily available alternatives are the most promising markers. To illustrate the principle in present-day writing, approval can be expressed across topics as ``great,'' ``good,'' ``terrific,'' or an emoji, whereas alternatives for ``cat'' stay tied to a topic and ``are'' has few. Synonym-based attribution dates to the eighteenth century \citep{koppel2011unsupervised}. A heuristic weighting words by their WordNet synonym count reached F1 0.310 on four nineteenth-century authors; combining stopword and common-name removal with global word- and synonym-frequency weighting raised F1 to 0.940 \citep{clark2007classifier, miller1995wordnet}. On the Hebrew Bible, clustering with the applicable subset of a 529-synset inventory separated Jeremiah and Ezekiel at 91\% chapter-level accuracy, a bag-of-words baseline performed near chance, and common-word representations seeded from the synset clusters then separated the documents almost perfectly with an SVM \citep{koppel2011unsupervised}. The texts are homogeneous and rich in content-dependent synonyms, which limits the result's transfer. Synonym normalization of common-word frequencies is the constructive use of the idea \citep{eder2022boosting}. As a standalone family the evidence is sparse and requires comparable opportunities for choice across texts.

\subsection{Complexity, vocabulary richness, and readability}\label{sec:complexity}

Word and sentence length are among the oldest measures in stylistic forensics \citep{mendenhall1887characteristic, morton1965authorship, mannion2004sentence}. Vocabulary richness includes vocabulary size, hapax and dis legomena, type--token ratio, and indices intended to reduce length dependence, such as Yule's K and Sichel's S \citep{yule2014statistical, tweedie1998variable}. Readability indices include Flesch--Kincaid, the Gunning fog index, and the Automated Readability Index \citep{kincaid1975derivation, gunning1952technique, smith1967automated}. Their use as identifiers is limited by length dependence and poor discrimination in several comparisons. The length-independence premise that justifies the richness indices is false for all of them \citep{tweedie1998variable}. \citet{hoover2003another} concludes that ``the basic assumption that it constitutes a wordprint for authors is false''. For the Pauline Epistles, Herdan called sentence length ``perhaps the most unsuitable of all available tests'' because original punctuation was absent \citep[p.~229]{morton1965authorship}, and Mendenhall thought word length useful only for eliminating candidates. In cross-language experiments with bilingual Reddit writers, the maximum macro-F1 reported for SVMs using the vocabulary-richness feature set labeled LIFE in that study was 0.110, against 0.479 for character n-grams across the tested conditions, which included machine translation \citep{murauer2021dt}. In a deception-detection study, one readability feature set reached 0.920 accuracy alone and 0.890 when added to a linguistic baseline; other combinations improved the baseline. This is evidence about fake-news classification, not authorship \citep{santos2020measuring}. Topic also affects these measures, as Yule's K and word lengths were among those that discriminated topic rather than author in Greek \citep{mikros2007investigating}. Complexity measures remain useful as supplementary descriptors inside broad feature inventories \citep{zheng2006framework}, including Writeprints \citep{abbasi2008writeprints}, where they sit beside word, punctuation, and grammatical features. A broad measure such as sentence length captures fewer of a writer's choices than a distribution of words or n-grams. A rewriting prompt that named only sentence length was nevertheless enough to disrupt attribution in short reviews \citep{khan2026iamnoone}, although the study did not test whether the edits were confined to sentence length, so a measure that is weak on its own is not thereby dispensable.

\subsection{Discourse, document structure, and graph measures}

Discourse-level features describe how a text is organized above the sentence. Entity-grid coherence patterns performed ``often as well as and sometimes better than standard features'' on 19 book-length texts by nine authors. The abstract makes no topic-independence claim. The topic robustness later attributed to discourse features was imported rather than tested there \citep{feng2014patterns}. In one chunk-length experiment, gains from discourse features appeared at roughly a thousand words for entity grids and six hundred for rhetorical-structure features \citep{wang2017leveraging}. Reasoning-structure graphs derived by argument mining extend this line to LLM-generated text and improve invariance at a cost in clean-text accuracy (Section~\ref{sec:llm-shift}) \citep{kikteva2026show}.

Structural features describe aspects of formatting such as openings and closings, paragraph length, indentation, trailing whitespace, or signatures. They help in registers that allow personalized arrangement such as e-mail and forum posts \citep{devel2001mining, zheng2006framework, abbasi2005applying} and for texts too short for general features \citep{stamatatos2009survey}. In de Vel et al.'s three-author e-mail ablation, they were worth ``only a maximum of a few percentage points''. In Zheng et al.'s five-author comparisons, adding structural features to lexical and syntactic features raised accuracy by 4.34--13.61 percentage points across the English and Chinese datasets and three classifiers \citep[Table~5]{zheng2006framework}. Because they are tied to register, their transfer across registers is limited.

Graph representations encode a text as a network of co-occurring words and read off degree, path length, centrality, clustering, assortativity, and burstiness \citep{mehri2012complex, antiqueira2007some, amancio2015complex}. Counts of thirteen directed three-node motifs gave results close to a twenty-most-frequent-word baseline with C4.5 and Naive Bayes, and worse with kNN and SVM \citep{marinho2016authorship}. The function-word adjacency network of Section~\ref{sec:fw} illustrates how these representations differ from frequency vectors. The audit discussed there shows why decision rules and corpus sizes also matter.

\subsection{Content words and topical shortcuts}\label{sec:content}

Content words (nouns, main verbs, adjectives, adverbs) carry a document's subject. They are often struck out as stylistic evidence altogether, although the evidence supports a narrower treatment.

The argument against them is a conditional-probability argument \citep{altakrori2021topic}. Suppose a user in a film-review corpus writes only about action films. A model trained on both content and function words will learn an association between that user and action-film vocabulary, possibly so strong that it learns nothing about the user's style that would transfer to a post on another subject. The model's within-corpus performance is inflated and its out-of-topic performance is not measured. Corpora in which topic and author are confounded by construction are the norm, not the exception, and it is prudent to assume the topic differs between training and test. Neural models that encode raw text are exposed to the same shortcut unless prevented \citep{geirhos2020shortcut}. Neural verifiers trained on PAN data are demonstrably ``biased towards named entities,'' with models trained after entity removal scoring better and transferring better to a dark-web forum corpus \citep{brad2022rethinking}.

Content-masking experiments help distinguish the effects of different word classes. On the Guardian10 corpus, masking all common nouns left the PPM character model's cross-topic accuracy nearly unchanged (70.41\% to 69.10\%), while masking only proper nouns raised it to 80.30\% across topics and from 69.84\% to 82.56\% across genres. On the single-domain IMDB1M corpus the same common-noun masking cost 14.3 points (64.77\% to 50.47\%) \citep{sundararajan2018represents}. A model built only from content-bearing words, TopicFit, reached AUC 0.95 even on topic-controlled subsamples of a PAN verification benchmark \citep{sawatphol2024addressing}. Content words carry word choice as well as topic (Section~\ref{sec:measures} on synonym choice), so a high score for a content-only model is a reason to check a benchmark for topic shortcuts rather than proof that it measures topic. The check is the topic-controlled rebuild of the split that \citet{sawatphol2024addressing} report. On the evidence above, proper nouns are the safest words to remove. The topic-neutrality of every other class is something to test on the corpus at hand rather than to assume. Exceptions exist where topic is aligned by construction, as in the Jeremiah--Ezekiel decomposition above \citep{koppel2011unsupervised}, but they are exceptions. What to do about the shortcut before or during representation learning, whether to mask the input, to control the training pairs, or to diversify the data, is the subject of Section~\ref{sec:masking}, and it is contested.

\subsection{Transfer as evidence for a measure}

Successful transfer is positive evidence for a measure, but it does not establish independence from content. Emoji features retained verification performance across subreddits in the studied population of emoji-using Reddit authors \citep{etli2026emoji}. Punctuation categories also performed well under the topic shifts tested by \citet{sapkota2015not}. These findings identify useful features for those comparisons; neither test isolates all possible content dependence. They support evaluating the shift a method must withstand, rather than assigning a family a permanent topic-independent label.

A single ranking of the families would be convenient. \citet{grieve2007quantitative}, the only study comparing 39 measurement types on one corpus, shows why a single ordering misleads. Rankings changed with the number of candidates. Its results hold for forty British columnists and no other setting. The typed n-gram inversion between CCAT-50 and Guardian1 \citep{sapkota2015not}, the different performance of word and character n-grams on translated texts \citep{bogdanova2014cross}, the emoji features that are weaker in-domain and more transferable out of it \citep{etli2026emoji}, and the Arabic survey finding that ``the effective features differ from one dataset to another'' \citep{alqahtani2022survey} point in the same direction. Table~\ref{tab:measures} therefore reports, for each family, what the in-domain evidence shows, what the shift evidence shows, and the failure mode, with the condition attached.

\begingroup
\tableformat
\begin{longtable}{L{0.14\linewidth}L{0.24\linewidth}L{0.29\linewidth}L{0.22\linewidth}}
\caption{Evidence summary for the measure families. Each cell summarizes the cited studies under the stated conditions; the failure-mode column gives the review's interpretation of that evidence, not a common-scale score. Absolute values from different rows are not directly comparable.}\label{tab:measures}\\
\toprule
\textbf{Family} & \textbf{In-domain evidence} & \textbf{Evidence under shift} & \textbf{Failure mode} \\
\midrule
\endfirsthead
\caption[]{(continued)}\\
\toprule
\textbf{Family} & \textbf{In-domain evidence} & \textbf{Evidence under shift} & \textbf{Failure mode} \\
\midrule
\endhead
Function words & Confirmed in many studies; 60 words give 81.5\% pairwise on 8 controlled authors \citep{baayen2002experiment} & Best of three families across catalog-defined domains \citep{menon2011domain}; insufficient across platforms \citep{overdorf2016blogs}; 4/10 frequent words carry topic \citep{mikros2007investigating}; genre sensitivity can outweigh author signal, as reviewed by \citet{nini2023theory} & Curated lists; residual topic and register sensitivity; distinct-symbol replacement reduced MNB accuracy \citep{markov2017improving} \\
Common words & 90\%/45\% at 2/40 candidates with ten-text cutoff \citep{grieve2007quantitative}; Delta optimum 1,000--2,000 most frequent words \citep{evert2015explaining} & Delta recovered original-author signal in literary translations \citep{rybicki2012great}; normalization is the decisive factor \citep{evert2017understanding} & Cutoff admits content; language- and genre-dependent optimum \citep{rybicki2011deeper} \\
Character n-grams & Top single family in 39-measure comparison \citep{grieve2007quantitative}; Highest or tied English-essay scores in AAAC \citep{juola2008authorship} & Best of four families cross-topic \citep{sapkota2014cross}; affix+punct suffice, best type inverts by condition \citep{sapkota2015not}; pruning to 341 features gives 73.2\% cross-topic \citep{markov2017improving}; default-parameter PPM is competitive cross-domain \citep{halvani2018crossdomain}; 95$\to$69 with language, feature and classifier changes \citep{bogdanova2014cross} & Stem-like n-grams carry topic; not language-robust \\
Punctuation & +16 pp on 40 columnists \citep{grieve2007quantitative}; top-10 feature at 100k blogs \citep{narayanan2012feasibility} & Best typed category cross-topic \citep{sapkota2015not}; transcript-version comparison does not isolate its contribution \citep{aggazzotti2025stylometric} & Editor confound \citep{darmon2021pull}; no dedicated transfer study \\
Syntax & Phrase-structure rules separate authors \citep{baayen1996outside}; syntax + character CNN 96.2 on IMDb62 \citep{zhang2018syntax} & up to +31 pp over lexis on topic-heterogeneous poetry, parity on topic-controlled news \citep{raghavan2010authorship}; loses 3/4 tasks vs.\ char LM \citep{sundararajan2018represents}; grammar model first on 7/12 corpora \citep{nini2026grammar}; Universal Dependencies layers with glossary pseudolemmas 95.6$\to$16.7 in cross-language translation matching \citep{cinkova2020stylometry} & Parser dependence on informal text; language-specific annotation values; pq-grams have no ablation \\
Idiosyncrasies & C4.5: errors + one other family beat lexical + POS, 68.8/71.8 vs.\ 61.7 \citep{koppel2003exploiting} & Emoji features nearly flat cross-subreddit while classical drop 26.6 pp \citep{etli2026emoji}; minimal correction costs 3.8 pp in native-language identification, an indirect result \citep{uluslu2026impact} & Sparse in short text; removed by editing \\
Synonym choice & 91\% on Jeremiah/Ezekiel using a subset of 529 synsets \citep{koppel2011unsupervised}; neighbor normalization 0.908$\to$0.959 \citep{eder2022boosting} & None located & Requires comparable opportunities; content-dependent \\
Complexity, richness, readability & Length-independence false \citep{tweedie1998variable}; one readability combination non-additive in deception detection \citep{santos2020measuring} & Cross-language SVM maxima: LIFE 0.110 macro-F1 vs.\ 0.479 for character n-grams, including translated conditions \citep{murauer2021dt}; Yule's K and word length discriminate topic \citep{mikros2007investigating}; a sentence-length-only rewriting prompt disrupts attribution \citep{khan2026iamnoone} & Weak alone; length-dependent \\
Discourse, structure, graphs & Entity grids competitive on 19 book-length works \citep{feng2014patterns}; adjacency-network encoding (Section~\ref{sec:representations}): 29\% mean error reduction, 38\% for two authors \citep{segarra2015authorship} & Discourse gains appeared at 600--1000 words in a chunk-length experiment \citep{wang2017leveraging}; argued canon-size confound in an early-modern-play word-adjacency-network application \citep{barber2020function} & Length thresholds; register-bound formatting \\
Content words & Content alone 0.795 vs.\ content-masked 0.786 \citep{zhu2021idiosyncratic}; common-noun masking $-$14.3 pp on single-domain IMDB1M \citep{sundararajan2018represents} & Proper-noun masking +10/+13 pp, common-noun masking 70.41$\to$69.10 on Guardian10 cross-topic \citep{sundararajan2018represents}; content-only AUC 0.95 on topic-controlled PAN subsamples \citep{sawatphol2024addressing} & Carry topic and word choice together; a high content-only score calls for a topic-shortcut check \\
\bottomrule
\end{longtable}
\endgroup
\section{Stylistic representations}\label{sec:representations}

A writing sample can be represented as a vector of explicit feature values (counts, frequencies, indices), as a learned embedding, or as a model of the author's text under which new text is scored, as in compression and per-author language models. This section reviews these representations in turn and then the scoring and decision rules applied to them. Across these approaches, changes to the encoding, aggregation, or scoring of measures can improve performance without enlarging the feature inventory.

\subsection{Feature vectors and their normalization}\label{sec:bow}

In a bag-of-words representation, each document is represented by the counts of its features, such as function words or character trigrams. Measures such as readability or mean sentence length contribute one value per document. This representation remains standard in stylometry and has sufficed for many historical-attribution problems with moderate candidate sets. Its limitation is that word order cannot be recovered beyond the local sequences captured by n-grams, while large $n$ brings sparsity and overfitting to phrases.

The same features yield different representations under different scaling. Relative frequencies put texts of different length on one scale; TF--IDF reweights features toward those concentrated in few documents. Word frequencies can reflect a community's topic, as the subreddit profiles studied by \citet{backes2015openworld} illustrate. For word-level features over a full vocabulary, this weighting favors topic words and is seldom useful in stylometry without modification, although character-level TF--IDF is a standard and competitive representation and the locally weighted variant of \citet{escalante2011local} improves on plain n-grams. In the Delta family, a sharper finding comes from crossing standardization and normalization with Manhattan, Euclidean, and cosine distances on 75 novels in each of three languages. The study reports that ``feature vector normalization, that is, the transformation of the feature vectors to a uniform length of 1 (implicit in the cosine measure), is the decisive factor'' in the improvement of recent Delta variants, and the author-identifying information ``lies in the profile of deviation across the most frequent words rather than in the extent of the deviation'' \citep{evert2017understanding}. Earlier mathematically better-motivated modifications of Delta ``do not yield better results'' \citep{evert2015explaining}. Length normalization removes a nuisance dimension into which text length and topic load. Normalization need not use the whole document as denominator, as the semantic-neighbor normalization of Section~\ref{sec:measures} shows \citep{eder2022boosting}. Function-word adjacency networks offer a related comparison with separately tuned word inventories and frequency baselines \citep{segarra2015authorship}. The feature inventory and its encoding are separate choices.

\subsection{Compression and generative language models}\label{sec:generative}

A second family represents an author not by a vector but by a model of their text under which new text is scored. PPM compression builds character-level context models without an explicit word vocabulary or a hand-selected feature inventory. It was ``strongest precisely on language and author identification'' and merely competitive on genre and topic \citep{teahan2003using}. Its cross-domain behavior was reviewed in Section~\ref{sec:charngrams} \citep{halvani2018crossdomain, sundararajan2018represents}. A common controlled comparison with per-author neural language models would help isolate what transfers (Section~\ref{sec:open}).

Neural generative representations inherit the idea, but separating author-specific from shared language patterns does not automatically remove domain dependence. A character-level RNN with a shared recurrent layer and one output head per author won the English and Greek PAN 2015 verification tasks but performed poorly on Dutch, which its author attributed to a large genre difference between reference and questioned texts \citep{bagnall2015author}. The choice of normalization data can help, as shown on a topic- and genre-controlled corpus where using unlabeled test texts for normalization let the best configuration of each of four language-model families beat a character-trigram SVM in all six genre-specific comparisons on 21 authors. The mean was 77.18\% across those four configurations (our calculation), against 45.05\% \citep{barlas2020cross}. The extension's mixed evidence on BERT layers \citep{barlas2021transfer} and a lower-performing reimplementation \citep{tyo2022state} make the normalization procedure and evaluation setting important parts of this result.

Per-author language models also offer a direct scoring rule: attribute a text to the model under which it has the lowest perplexity. Fine-tuned GPT-2 models improved on n-gram and compression baselines on Blogs50, but the n-gram baseline remained better on CCAT50; these were within-domain comparisons \citep{huang2024alms}. Fine-tuned encoders such as BertAA instead learn a discriminative classifier \citep{fabien2020bertaa}, while multi-source systems combine lexical, syntactic, and embedding features with scenario-specific selection \citep{custodio2021stacked}. These approaches differ in what they model and how they use the reference writing, not merely in benchmark rank.

\subsection{From word embeddings to text representations}\label{sec:embeddings}

Word embeddings represent words as dense vectors trained on the distributional hypothesis \citep{harris1954distributional, firth1952synopsis, mikolov2013efficient, pennington2014glove}. However, the use of dense vectors in place of discrete features does not by itself improve authorship identification. Comparisons with word2vec and GloVe found that results varied across languages, with no consistent advantage over character-based models \citep{rahgouy2019cross, ruder2016character, shrestha2017convolutional}.

The choice of what to embed matters, as information beyond individual word meanings can be retained through continuous n-gram representations \citep{sari2017continuous}, a POS-tag channel alongside GloVe \citep{jafariakinabad2019style}, or DeepStyle's combined word and character inputs \citep{hu2020deepstyle}. For instance, doc2vec trained on character, word, and POS n-grams outperformed word embeddings and character-n-gram linear models on the Guardian cross-topic corpus, an advantage that the authors attributed to the n-gram inputs \citep{gomezadorno2018document}. This comparison connects learned representations to the feature choices reviewed in Section~\ref{sec:measures}.

Contextual embeddings from ELMo and BERT-family encoders \citep{peters2018elmo, devlin2018bert, liu2019roberta} give each token a context-dependent vector and are the starting point for most learned author representations \citep{zhu2021idiosyncratic}. The choice of layer matters because lower layers, which mainly carry syntactic information, outperform higher, more semantic layers for author identification in the experiments of \citet{jafariakinabad2021unifying}, whereas the layer comparison in \citet{barlas2021transfer} is mixed. In another comparison, stylistic features are equally readable at every layer of authorship-trained encoders and an off-the-shelf encoder never trained on authorship \citep{kulumba2026where}. Generic embedding geometry is topic-first \citep{icard2025embedding}, so a token representation is not yet a style representation or a representation of a document or an author. Token vectors are pooled into a passage vector, then passage vectors are combined across documents. The representation depends on the encoder as well as the sampling and aggregation of the available text.

\subsection{Learned author and style representations}\label{sec:learned}

The training objective determines which similarities a learned representation preserves. Table~\ref{tab:learned} summarizes the main lines by training signal and by the mechanism, if any, through which content is controlled.

\paragraph{Author-supervised metric learning} In author-supervised contrastive learning, texts or collections by the same author are positives and those by different authors negatives. Training brings positives together relative to negatives, allowing the representation to be used for authors unseen in training. Early account-linking models learned embeddings of social-media activity from account identities and accommodated variable-sized samples and open-set evaluation \citep{andrews2019learning, khan2021deep}. LUAR makes the collection-level representation explicit by encoding excerpts and aggregating them by attention to produce one vector for an author's writings \citep{riverasoto2021learning}.

A related line uses Siamese networks to compare texts, including AdHominem's hierarchical attention model \citep{boenninghoff2019explainable} and architectures for large and one-shot candidate sets \citep{saedi2021siamese}. AdHominem's successors combine topic handling with probabilistic scoring and, later, an option to abstain \citep{boenninghoff2020deep, boenninghoff2021o2d2}. Their task-specific scores are recorded in Table~\ref{tab:readouts}. Representation learning, scoring, and abstention are designed together. Contrastive fine-tuning can also improve closed-set attribution, although gains need not extend to every author \citep{ai2022whodunit}. PART pretrains a transformer representation for authorship \citep{huertastato2022part}. These methods learn from author labels without requiring an explicit stylistic inventory. Whether those labels teach style or correlated content is the next question.

\paragraph{Controlling content through the training pairs} If different-author examples come from unrelated discussions, topic separates them and the model can learn topic. \citet{wegmann2022same} sampled different-author examples from the same conversation, the same domain, or unrestricted data. Testing an uncontrolled model with same-conversation negatives exposed its reliance on topic, as the AUC drop in Section~\ref{sec:factors-topic} shows. Training with these negatives keeps the original words while discouraging reliance on content differences. Topic-regularized learning uses TF--IDF similarity to rescale embedding similarities before distillation \citep{sawatphol2022topic}. A content-negative contrastive objective raised Blog attribution from 0.3928 to 0.4469 over simple fine-tuning \citep{huertastato2024isolating}. Probabilistic content masking combined with single-language batches serves the multilingual case \citep{kim2025leveraging}. A variational information bottleneck with a stylistic constraint, in which the document embedding must predict a predefined 300-feature vector including function-word, punctuation, POS, and readability measures, learns joint author and document embeddings whose individual dimensions correlate with named features and whose accuracy--fidelity trade-off is explicit \citep{terreau2024capturing}. A topic-debiasing attention mechanism that reweights tokens by their loading on a latent topic dictionary is the direct neural instance of ``debias rather than mask,'' motivated by the concern that learned stylometric representations can retain topical bias \citep{hu2023tdrlm}. Its ancestor learned four modality-specific stylometric representations with a jointly trained topical--lexical pair intended to separate topic from word choice. Its authors note that ``for short text its performance will degrade'' and that performance varies between formal and informal text, so domain dependence survived the disentangling \citep{ding2019learning}.

\paragraph{Specifying the stylistic distinction directly} Training examples can encode the intended distinction instead of relying on author identity. StyleDistance generates parallel examples in which one feature (contraction use, formality) varies while meaning is held, arranges them into triplets whose negative differs in style and shares content, and combines synthetic with natural examples \citep{patel2025styledistance}. mStyleDistance extends the inventory across languages and adds cross-language triplets \citep{qiu2025mstyledistance}. A style encoder can also be trained with no labels of either kind by \emph{paraphrase inversion}, which reconstructs human text from its machine paraphrase with a frozen semantic encoder. This architecture is intended to leave the style encoder only the non-semantic residual. Trained on 63,184 Reddit authors, the resulting representation transfers zero-shot to authorship verification at AUROC 0.74 against LUAR's 0.78 and to the STEL style-discrimination task (matching sentence pairs by attributes such as formality or complexity) at 84\% accuracy against StyleDistance's 85\%. In a separate multiple-target few-shot machine-text detection experiment with original and paraphrased support examples, it stays above 0.90 AUROC(1), the source's partial-area score restricted to false-positive rates up to 1\%, with 80\% of query documents paraphrased \citep{riverasoto2026unsupervised}. By operationalizing style as the residual needed to reconstruct a text from its paraphrase, this design uses the rewriting operators of Section~\ref{sec:factors-mediation} as a training signal.

\paragraph{Interpretable dimensions} LISA represents a text by predicted agreement with natural-language descriptions of stylistic attributes, trained on LLM-generated annotations \citep{patel2023lisa}. NeuroBiber predicts the presence of predefined Biber-style features from a rule-based tagger's labels, predicting presence rather than counts \citep{alkiek2025neurobiber}. IDIOLEX offers unified continuous representations of idiolectal and stylistic variation \citep{kantharuban2026idiolex}. Post-hoc natural-language style description generated directly from text is ``prone to the LLM's biases and hallucinations'' \citep{kim2026interpreting}. The alternative curates 1,010 style features in 26 categories, generates text conditioned on them, and trains a decoder to recover the eliciting prompt from a representation. Prompt recovery is evaluated separately from whether the recovered prompts regenerate matching style. Activation-steering analysis finds that per-aspect style directions ``share a common authorship backbone while conflicting on aspect-specific residuals,'' which explains why naive merging of style dimensions fails while offering a structural hypothesis for uneven transfer across registers \citep{tran2026can}. These lines connect learned representations to the lexical and grammatical choices of Section~\ref{sec:measures} while differing in how features and labels are obtained.

\subsection{What learned representations encode}\label{sec:probe}

Early studies reported that attribution accuracy is not by itself evidence that a representation captures style. Recent author-embedding models ``underperform simple baselines built on pretrained sentence embeddings'' when evaluated by how well stylistic features can be predicted from them, and are ``mostly driven by the inner semantic of authors' production'' \citep{terreau2021writing}. Two years later, targeted comparisons involving masking, paraphrasing, and style or topic prediction found that authorship representation learning captures stylistic features only loosely \citep{wang2023authorship}. The two properties can be measured separately. On the STEL-or-Content probe, which asks whether a representation places a text nearer to a same-style or a same-content alternative, the contrastively trained model without content control scores 0.240 and its conversation-controlled counterpart scores 0.420 \citep{wegmann2022same}. On STEB, LUAR tops the retrieval task (77.64; R@1 83.0, mean reciprocal rank (MRR) 87.8, both reported on a 0--100 scale) with a content-independence score of 0.106, while CISR, the content-controlled representation trained using same-conversation negatives, scores 0.323 on content independence and 45.87 on retrieval \citep{riverasoto2026steb}. In that benchmark no model scored well on both. STEL-or-Content is a binary choice with a random baseline of 0.500. The overall scores reported by \citet{wegmann2022same} are below that baseline: a representation that favors the same-content alternative over the same-style alternative fails the test. Conversely, a high score can reflect avoidance of shared content without adequate style discrimination, so the authors recommend interpreting it alongside authorship verification and the original STEL task.

Controlled comparisons show how much a learned representation's performance can depend on content. On Amazon, retraining a Siamese RoBERTa on content-masked text costs 0.045 F1 (0.831 to 0.786), and the same representation predicts product domain worse than a fine-tuned RoBERTa but above chance \citep{zhu2021idiosyncratic}. This small loss supports the presence of a style signal beneath the content. With content controlled exactly by construction, parallel Bible translations with a tunable content--style association, a RoBERTa classifier trained where content perfectly predicts the label scores 94.9\% when evaluated matched and 55.3\% when content ceases to predict, against 50\% chance, so 88.2\% of its above-chance margin was lost (our calculation), while a model trained where content cannot help scores 87.4\% and 86.7\%, with little change \citep{liu2026style}. Probing a fine-tuned Llama-3 8B on 27 literary authors finds that author identity rides on word order, pronoun use, and contextual word choice, that punctuation and capitalization separate authors but carry no genre-level signal, and that the model partly exploits memorized pretraining text, a reason to check memorization as an alternative explanation when evaluating pretrained LLMs on public texts \citep{hicke2025looking}.

Scale may trade against robustness. Paraphrase sensitivity grows monotonically with training-corpus size. The paraphrase-induced MRR drop is 0.169 at 100,000 training authors, 0.237 at one million, and 0.261 at five million, with paired tests significant at each scale \citep{wang2023authorship}. This is a single study, and paraphrase changes wording as well as content, so the trend shows growing sensitivity to paraphrase rather than a measured increase in topic reliance.

\subsection{Scoring and decision rules}\label{sec:readouts}

Table~\ref{tab:readouts} separates changes to the scoring rule from changes that also involve training or the evaluation task. Scores within a row describe the cited comparison, not a ranking across rows.

Several classical methods seek robustness through their scoring and decision rules. In learned systems, the scoring mechanism also shapes the representation. \citet{kulumba2026where} fine-tuned models from the same pretrained backbone with the same data and the same loss, differing only in the scoring mechanism through which authorship was read out during training and at inference; the model trained with mean pooling reached R@20 of 0.121 and the model trained with patch-level late interaction 0.497. At the same time, the stylistic features that stylometry relies on (word length, punctuation density, function-word frequency) were equally readable by linear probes at every layer of every model, including an off-the-shelf general-purpose encoder never trained on authorship, and the authors conclude that ``the main question is not, therefore, which model encodes more stylistic information, but which scoring mechanism can effectively read it out.''

The impostors method makes the decision relative to a background population sampled with random feature subsets \citep{koppel2014determining}. Selecting impostors by highest similarity to the known documents (hard negatives) and scoring by rank rather than by a binary criterion raised AUC most on the cross-genre PAN 2015 sets (Greek 0.66 to 0.84, Spanish 0.79 to 0.85) and slightly lowered it on English \citep{potha2017improved}. A later correction estimates domain-specific score differences to mitigate domain effects in the vector representation \citep{khonji2023improved}. However, under cross-fandom open-set conditions the impostors baseline was the worst of the three PAN 2019 baselines (macro-F1 0.395 against 0.545 for an SVM and 0.533 for a compressor), and no submission that year used deep learning \citep{kestemont2019overview}. Four binary-extrinsic systems were classified as non-deterministic because their procedures involve randomness \citep{halvani2019assessing}.

Unmasking reads a degradation curve rather than a distance and asserted independence of ``language, period and genre'' on the strength of one Hebrew--Aramaic case \citep{koppel2004authorship}. In a cross-genre evaluation, it fell from 96\% to 77\% with genre- and author-specific parameters \citep{kestemont2012evaluating}. Explicit topic models can be used as calibration devices inside an impostor pool even where topic is a confound as a feature \citep{potha2019improving}.

Aggregation and abstention also affect the decision. At 10,000 candidates a robustness filter turned 46\% accuracy into 93.2\% precision at 39.3\% recall \citep{koppel2013authorship}. The one PAN system whose score rose between the closed-set and open-set years had added abstention through out-of-distribution detection, among other changes \citep{boenninghoff2021o2d2}. Framing changes the number reported for one system on one dataset (Section~\ref{sec:protocol-effects}). Verification methods with fixed decision criteria are not comparable with threshold-tuned ones. Many fail once required to accept the true author's documents and reject others simultaneously \citep{halvani2019unary, halvani2018rethinking}.

Prompted language models can also make authorship decisions. Step-by-step stylometric prompting uses a classical feature inventory to guide the model's reasoning \citep{hung2023who}. On the same ten-candidate task, \citet{huang2024large} found that linguistically informed prompting moved one model from Micro-F1 36.67 to 86.67, a 50-point effect of wording alone measured over three repetitions. Increasing the candidate count from 10 to 20 cost 24 points. One-shot style-transfer log-probabilities provide another, likelihood-based score \citep{mirallesgonzalez2025shot}. Fixed style embeddings and a GPT-4o judge perform differently across registers. On 600 human/LLM continuation pairs across six domains, the embeddings reached 81.5\% and the judge 67.7\% on GPT continuations. The judge performed better on fiction and academic text, while the embeddings performed better on spoken transcripts and scripted dialogue \citep{abbas2025attribution}. Under a calibrated likelihood-ratio approach, a classical stylometric system and an embedding system are complementary rather than substitutes, with fusion giving the best log-likelihood-ratio cost ($C_{llr}$) of 0.325 on Japanese blog text \citep{ghatpande2026fusing}.

\begingroup
\tableformat
\begin{longtable}{L{0.39\linewidth}L{0.13\linewidth}L{0.22\linewidth}L{0.17\linewidth}}
\caption{Comparisons of scoring and decision rules. The first two rows change training or selection as well as scoring; the third compares successor systems on different tasks, not an isolated abstention effect. Values belong to the stated comparison and should not be ranked across rows.}\label{tab:readouts}\\
\toprule
\textbf{Contrast} & \textbf{Metric} & \textbf{Baseline $\to$ changed} & \textbf{Source} \\
\midrule
\endfirsthead
\caption[]{(continued)}\\
\toprule
\textbf{Contrast} & \textbf{Metric} & \textbf{Baseline $\to$ changed} & \textbf{Source} \\
\midrule
\endhead
Mean pooling vs.\ patch-level late interaction, models fine-tuned separately from one backbone & R@20 & 0.121$\to$0.497 & \citet{kulumba2026where} \\
Random impostors vs.\ hard impostors with rank scoring (two changes), PAN15 Greek / Spanish / English & AUC & 0.66$\to$0.84 / 0.79$\to$0.85 / decrease $<$0.01 & \citet{potha2017improved} \\
PAN 2020 closed set vs.\ PAN 2021 open set, successor system with out-of-distribution abstention and other changes (different tasks) & AUC & 0.97$\to$0.99 & \citet{boenninghoff2021o2d2} \\
No guidance vs.\ linguistically informed prompt, GPT-4 Turbo, Blog, 10 candidates, three repetitions & Micro-F1 & 36.67$\to$86.67 & \citet{huang2024large} \\
\bottomrule
\end{longtable}
\endgroup

\subsection{Structured and alternative representations}\label{sec:other}

Representations borrowed from other fields provide occasional additional signal. Chaos-game encoding of characters into a two-dimensional image, developed for DNA and also illustrated on English \citep{jeffrey1990chaos}, allows image classifiers to be applied, at the cost that sixteen base-4 codes cannot distinguish all characters (``g,'' ``h,'' and ``j'' share a code) \citep{stoean2020author}. Word co-occurrence graphs and their motif counts were reviewed in Section~\ref{sec:measures} \citep{mehri2012complex, marinho2016authorship}. Dependency trees are graphs and local tree structure has been modeled as such \citep{sidorov2013syntactic, tschuggnall2014enhancing, murauer2021dt}. Argument-mining reasoning graphs classified by a GNN represent a text above the lexical level. For attributing LLM-generated essays, they beat text baselines by up to 27 points under paraphrase and French or Turkish back-translation and by 19 points on unseen model versions, while being worse on clean text (0.660--0.810 against 0.900 accuracy) \citep{kikteva2026show}. Whether this explicit accuracy-for-invariance trade holds for human authors has not been tested. In general these representations are most useful in ensembles beside established features.

\begingroup
\tableformat
\begin{longtable}{L{0.17\linewidth}L{0.2\linewidth}L{0.24\linewidth}L{0.3\linewidth}}
\caption{Stylistic representations by encoding or training signal and content-control mechanism. The final column distinguishes shift tests, probing tasks, and other evaluations; absence of an explicit content control does not establish content independence. Values belong to their cited studies, not a common benchmark.}\label{tab:learned}\\
\toprule
\textbf{Representation} & \textbf{Encoding or training signal} & \textbf{Content control} & \textbf{Evaluation; note} \\
\midrule
\endfirsthead
\caption[]{(continued)}\\
\toprule
\textbf{Representation} & \textbf{Encoding or training signal} & \textbf{Content control} & \textbf{Evaluation; note} \\
\midrule
\endhead
Frequency vectors / Delta \citep{evert2017understanding} & Standardized and normalized frequent-word counts & No explicit content-removal step & Comparison on novels in three languages, not cross-language transfer \\
Function-word adjacency networks \citep{segarra2015authorship} & Transition relations between function words & Adaptively selected function words & Separately tuned frequency baselines; topic independence not tested \\
Per-author language models \citep{barlas2020cross, barlas2021transfer} & Sequence prediction with shared and author-specific components & Normalization corpus & Cross-topic and cross-genre attribution; normalization data affect results \\
Invariant user embedding \citep{andrews2019learning} & Metric learning with account-identity proxy labels & None & Time (behavior drift); unseen users; Reddit, Twitter, Wikipedia \\
Account linking \citep{khan2021deep} & Account-identity supervision; variable-size activity samples & Topic-matched linking test & Unseen accounts; open-set protocol \\
LUAR \citep{riverasoto2021learning} & Author-supervised contrastive, attention pooling & None & Cross-domain Reddit/Amazon/fan fiction; retrieval \\
AdHominem / DBFS / O2D2 \citep{boenninghoff2019explainable, boenninghoff2020deep, boenninghoff2021o2d2} & Siamese with attention; PLDA Bayes factor & Topic masking; fandom prefix; OOD abstention & Cross-fandom, closed set (2020) and open set (2021) \\
Contra-X \citep{ai2022whodunit} & Contrastive fine-tuning, closed set & No dedicated mechanism; shared prompts in machine-text corpus & Blog/IMDb attribution; machine-text authors also tested; per-author trade-off \\
Same-conversation negatives \citep{wegmann2022same} & Author-supervised contrastive & Negatives share conversation/domain & Reddit; uncontrolled model AUC 0.79$\to$0.58 when test negatives share conversation \\
Topic-regularized \citep{sawatphol2022topic} & Distillation of author embeddings & TF--IDF-based similarity rescaling & Cross-topic \\
Style--content disentanglement \citep{huertastato2024isolating} & Contrastive fine-tuning of STAR & Negatives from a frozen content encoder & Blog, +0.054 \\
Multilingual author representations \citep{kim2025leveraging} & Contrastive & Probabilistic content masking; single-language batches & Unseen languages and domains \\
VADES \citep{terreau2024capturing} & Author--document pair learning + variational information bottleneck & Stylistic-feature constraint & Gutenberg, Blog, IMDb62; dimensions map to features \\
TDRLM \citep{hu2023tdrlm} & Verification & Topic-debiasing attention & Twitter--Foursquare, 3 tweets/sample, AUC 0.92 validation / 0.91 test \\
StyleDistance / mStyleDistance \citep{patel2025styledistance, qiu2025mstyledistance} & Contrastive triplets, synthetic and natural text & Negative shares content with anchor or positive, differs in style & STEL-style tasks; multilingual \\
Paraphrase inversion \citep{riverasoto2026unsupervised} & Reconstruct original from paraphrase, frozen semantics & Architectural bottleneck & Unseen LLMs; zero-shot verification AUROC 0.74 vs.\ LUAR 0.78; multiple-target few-shot machine-text test, original + paraphrased support, AUROC(1) $>$0.90 at 80\% paraphrased queries \\
LISA / NeuroBiber \citep{patel2023lisa, alkiek2025neurobiber} & Predict named attributes & Interpretable dimensions & Attribute prediction \\
Style-eliciting prompt inversion \citep{kim2026interpreting} & Decode prompt from representation & 1,010 features, 26 categories & Prompt recovery and style-control regeneration \\*
Activation steering A3S \citep{tran2026can} & Training-free contrastive prompting & Backbone + residual decomposition & GPT-4.1-judged preference over TinyStyler on LaMP/LongLaMP \\
\bottomrule
\end{longtable}
\endgroup
\section{What the cross-domain evidence tells us}\label{sec:generalization}

Performance falls when the factors of Section~\ref{sec:factors} differ between reference and test, sometimes to chance \citep{wang2021cross}. Because a stylometric profile is estimated from a partial sample of a writer's output, it inherits the sample's biases. Most of the work reviewed here tries to isolate stylistic variation from the background variation that topic, genre, register, mode, time, language, and editing contribute. The contrasts in Table~\ref{tab:generalization} support three judgments, developed below: a benchmark score can be interpreted only after inspecting what the benchmark separates; an intervention has to be judged by the information it preserves as well as the topic it removes; and rewriting can be used to test which identifying signals a representation depends on.

\subsection{A benchmark's labels do not describe what it varies}\label{sec:benchmarks}

The first lesson is that ``cross-topic'' and ``cross-domain'' are properties of a split that must be checked, not read off its name. The PAN cross-domain verification tasks defined domain as fandom, a blend of topic and sub-genre \citep{kestemont2020overview, kestemont2018overview}. Within those tasks, same-author pairs are more topically similar than different-author pairs and the misclassified different-author pairs skew to higher topic similarity \citep{kestemont2021overview}, which is in-benchmark evidence of a residual shortcut beside the near-ceiling result on unseen fandoms noted in Section~\ref{sec:identifier}. On random subsamples of a cross-topic PAN benchmark a content-only model, TopicFit, reached AUC 0.96 and an overall score of 0.874, and the top train--test topic similarities inside the splits were 0.971 and 0.961 \citep{sawatphol2024addressing}. When the split was rebuilt to control topic similarity, the content-only model lost significantly (0.874 to 0.852) while the character n-gram model scored 0.927 and 0.926, with no statistically significant difference reported.

Neural verifiers trained on PAN data are ``biased towards named entities,'' and five new splits designed to separate topic from style, with entity removal, gave models that scored better and transferred better to dark-web forum data \citep{brad2022rethinking, manolache2022veridark}. The community's shared evaluation ground for human cross-domain verification then disappeared when PAN retired classic verification after 2023 in favor of generative-AI verification and multi-author style analysis \citep{bevendorff2024overview, bevendorff2026overview}, at the point when representation-learning methods were becoming standard. The 2025 multi-author task is the best topic-controlled shared task located. It evaluates sentence-level style-change detection on Reddit with an ``easy'' tier of varied topics and a ``hard'' tier of single-topic documents with high stylistic similarity. The best system's average F1 across tiers is 0.870 against 0.959 on the easy tier alone. In the parallel generative-AI track only 7 of 20 systems beat a TF--IDF SVM baseline \citep{bevendorff2025overview}. Resources built to control topic by construction exist \citep{halvani2016reddit, riddell2021call, wang2022cctaa}. On a controlled cross-topic Chinese newswire split both a RoBERTa classifier and a function-character-n-gram SVM ``perform below expectations'' \citep{wang2022cctaa}. The HIATUS HRS corpora illustrate how to separate genre from author identity through a within-author manipulation. They contain 855 foreground documents by recruited authors who each wrote in all five genres (board-game reviews, instructions, citizen journalism, literature forum posts, STEM forum posts). Their cross-genre protocol specifies that ``for every author, the needle documents were drawn from all genres except the query genre'' \citep{iarpa2024hiatus}. The design is documented in the program data sheet; a fuller methodological account would make its controls easier to assess. Changing the author population along with genre does not provide the same comparison.

The largest uniform evaluation reinforces the point from the other direction. AuthBench evaluates 47 neural models and three non-neural baselines zero-shot on 428,150 documents by 153,825 authors in ten languages, 66 fine-grained genres, and four length buckets. The best retrieval model reaches Success@5 of 0.258, while the best verification model reaches an equal error rate (EER) of 0.076. Different model families lead the two tasks, with performance differing widely across languages, genres, and lengths \citep{huang2026authbench}. Near-ceiling AUC on unseen fandoms within one platform and one genre \citep{kestemont2021overview} and Success@5 of 0.258 across languages and genres are not in conflict. They measure transfer across fandoms within one platform and transfer across languages and genres, respectively.

The reviewed studies do not provide a common author-from-topic-only or author-from-metadata-only baseline across the standard attribution corpora. The named-entity probe above is a model-side diagnostic. This is listed in Section~\ref{sec:open}.

\subsection{Masking removes author signal as well as topic}\label{sec:masking}

Masking should be judged by whether it reduces reliance on topic while preserving useful author differences. Removing content through masking, distortion, or preprocessing can affect both. The results run in both directions, and the studies that report gains and those that report losses differ in model class, in what was masked, and in task, so the direction cannot yet be attributed to any one of these.

For shallow frequency and n-gram models under an explicit topic shift, masking can help. Text distortion, which replaces less frequent words with a placeholder while preserving punctuation, capitalization, and frequent tokens, gained 8.9 points cross-topic (66.2 to 75.1) and 19.5 cross-genre (39.4 to 58.9) on the Guardian corpus at a cost of 2.4 points within topic on CCAT-10 \citep{stamatatos2017authorship, stamatatos2018masking}. Digit and named-entity replacement plus punctuation splitting gained 4.4 points with SVM and 5.9 with MNB cross-topic. On single-topic CCAT-10, digit replacement alone gained at most 0.8 points while the full bundle reduced accuracy \citep{markov2017improving}. Masking proper nouns raised a character language model by about 10 points cross-topic and 13 cross-genre, while masking all common nouns left cross-topic accuracy nearly unchanged and cost 14.3 points on a single-domain corpus \citep{sundararajan2018represents}. POS-noise pre-processing improved a grammar model on all twelve of twelve corpora, by 0.005 to 0.080 \citep{nini2026grammar}.

For learned contrastive representations under fixed resource-based masking, the studies below report losses. POS-based content masking made cross-genre retrieval worse (Success@8, the fraction of queries with a correct match in the top eight, 0.225 to 0.199) and per-genre retrieval worse still (0.570 to 0.459) \citep{fincke2024separating}. Stopword masking and POS masking underperformed a no-masking baseline (R@8, reported on a 0--100 scale, 10.34 and 19.14 against 22.60), a result the authors call ``surprising,'' noting that such schemes depend on ``high-quality, language-specific resources,'' while a learned probabilistic content mask improved on the baseline (24.66) \citep{kim2025leveraging}. POSNoise's own table sits below the unmasked original in most cells, although its published claim is that it beats text distortion, not that it beats no masking \citep{halvani2021posnoise}. On a Siamese encoder, masking every content word costs 0.045 F1 \citep{zhu2021idiosyncratic}.

Across these studies, gains appear for shallow models under an explicit topic shift and losses for learned representations under fixed resource-based masks, with the one learned masking scheme improving on its baseline. On this pattern, what a mask removes and how the model uses what remains determine the result together, so neither can be assessed alone. Section~\ref{sec:open} lists the experiment that would separate them. One hypothesis is that the outcome depends on how much author-specific word choice the mask removes relative to topical cues. Both frequency models and learned encoders can use that word choice. Crossing mask type with model class would test whether either is more sensitive to its removal. Masking also has a cost independent of performance because it removes the synonym choices and some idiosyncrasies of Section~\ref{sec:measures}. The alternatives are to control content in the training pairs rather than in the input \citep{wegmann2022same, huertastato2024isolating}, to model topic explicitly and factor it out \citep{potha2019improving, hu2023tdrlm}, to learn the residual needed beyond a frozen semantic representation \citep{riverasoto2026unsupervised}, or to curate the training data, which is the next lesson.

\subsection{The reference collection is part of the method}\label{sec:diversity}

A representation learned from a narrow sample of a writer's output can depend on that sample's subject matter. Experimental, correlational, and system-level studies show why reference-collection diversity belongs in the method design. In an experimental comparison, multiple-cross-topic training exceeded within-topic training in several genre cells, with character n-grams gaining most, although those pooled cells also use more training data (Section~\ref{sec:factors-topic}) \citep{sapkota2014cross}. The correlational result is that an author's number of training fandoms correlated with that author's test F1 in PAN 2018 \citep{kestemont2018overview}. The system result is that on the HIATUS cross-genre sets a retrieve-and-rerank pipeline whose reranker was trained with a curation scheme designed to avoid topical cues gained 22.3 and 34.4 absolute Success@8 points over prior systems on HRS1 and HRS2 \citep{agarwal2025crossgenre}. The gain belongs to the whole pipeline. Its design follows the same group's finding that standard information-retrieval training recipes are ``fundamentally misaligned with cross-genre'' attribution because retrieval objectives reward topical matching \citep{fincke2024separating}. Taken together they show that the composition of the training data affects transfer and deserves to be evaluated alongside the model, but do not rank it against other interventions. In practice, an analyst with topically diverse reference samples is in a better position than same-topic accuracy figures alone would suggest.

Domain-adaptation machinery was uncommon in the literature retrieved for this review. One example is domain-adversarial training of writing-style representations, in the domain-swap study whose zero-knowledge collapse was reported in Section~\ref{sec:factors-genre} \citep{bischoff2020importance}. A targeted search for adversarial and gradient-reversal adaptation on 17 September 2026 retrieved no additional authorship application. The approaches reviewed above pursue invariance through feature deletion, pair construction, and data curation.

\subsection{Rankings are protocol-dependent}\label{sec:protocol-effects}

Protocol choices reverse published rankings on the same data. In STEB's multilingual verification comparison, MSR, the multilingual authorship embedding model of \citet{kim2025leveraging}, ranks first under fixed truncation. Under chunk-and-pool aggregation StyleDistance and CISR rank first and second. The benchmark's authors write that ``without them our conclusions would differ materially'' \citep{riverasoto2026steb}. Standardizing text length and dataset size changes a drift benchmark's own model ranking while its genre-difficulty ordering survives \citep{kiefer2026when}. Which of two methods wins can reverse across randomly selected topic subsets, while topic-controlled splits stabilize the ranking \citep{sawatphol2024addressing}. Metric choice also changes the reported result for the same system and data (Table~\ref{tab:generalization}) \citep{tyo2022state}. Prompt wording moved an LLM by 50 Micro-F1 points \citep{huang2024large}. The PAN composite of four metrics has been criticized as an optimization target \citep{kestemont2020overview, tyo2022state}.

Reimplementing fifteen influential author-identification papers from scratch, none of which had an open implementation, was itself a research contribution \citep{potthast2016who}. Reproduction fails even with code and data because of unfixed test sets and homogeneous corpora \citep{riddell2021call}. Validation, in the forensic view, attaches to a protocol for a specific case question and not to a system \citep{grant2022idea}. Any performance figure should carry its task, corpus, candidate count, text length, shift axis, and metric.

Rankings reverse between benchmarks as well as between protocols. A logistic regression over Koppel-style features, an SVM over the Writeprints-static inventory, and a fine-tuned RoBERTa reached 0.640, 0.690, and 0.870 on the same-topic Extended Brennan--Greenstadt corpus of 45 authors, and 0.290, 0.290, and 0.190 on the cross-topic Riddell--Juola corpus, so the system that ranked first on one benchmark ranked below both feature-based systems on the other \citep{wang2025dissertation}.

\subsection{Rewriting as a probe of what a representation preserves}\label{sec:llm-shift}

Section~\ref{sec:factors-mediation} gave the graded effect of editing. Here we ask which representations survive rewriting or a change from human to model authorship. The evidence comes from three tasks, machine-text detection, individual authorship verification, and attribution of machine text to its generating model. Each paragraph names the task its evidence comes from.

\paragraph{What survives rewriting} In machine-text detection, discourse-level paraphrase with DIPPER drops a likelihood-based detector from 70.3\% to 4.6\% at 1\% false-positive rate. The proposed defense is retrieval against a store of previously generated text rather than any representation of the paraphrase \citep{krishna2023paraphrasing}. In individual authorship verification, off-the-shelf LLM paraphrasers achieve obfuscation success rates of 83\% and 92\% against a BigBird verifier and targeted impersonation of 78\% on fan fiction, so a transformer verifier that performs well in distribution can be steered as well as evaded \citep{alperin2025masks}. Aggregation changes the outcome in machine-text detection. Detectors that operate in a stylistic feature space hold about 0.95 AUROC against generators optimized to fool likelihood-based detectors. Single-document attacks succeed universally, but with five or more samples per author the distributions re-separate. Only a paraphraser optimized in the style-feature space itself closes the gap, with aggregation over documents restoring robustness \citep{riverasoto2025attacks}. GPT-4o impersonations of authors across e-mails, text messages, and social posts ``failed to sufficiently replicate authorial individuality to bypass established AV systems'' including AdHominem and LUAR. Some methods were more accurate on impersonations than on ordinary negative pairs because LLM output has greater lexical diversity than human writing \citep{zeng2026impersonation}. Under a forensic likelihood-ratio framework, ChatGPT imitations of 300 Amazon reviewers were evidentially null ($C_{llr}\approx0.967$--1.002 against 0.677 for the human baseline) \citep{ishihara2026chatgpt}. These results differ according to whether the rewriting was naively prompted or optimized against the target style.

Rewriting studies are most informative when they identify which changes disrupt a representation. In individual authorship attribution, a sentence-length prompt disrupted identification nearly as much as a full style-profile prompt \citep{khan2026iamnoone}. This suggests a useful intervention to test, but the study did not isolate sentence-length changes from the other edits produced. Evidence from representation probes \citep{hicke2025looking}, translated-text comparisons \citep{rybicki2025can}, and LLM restyling \citep{icard2026measuring} identifies additional cues and model differences, but does not establish which cue caused a rewriting attack to succeed.

Moving above lexical features offers another possibility. For attribution of machine text to its generating model, reasoning graphs improved performance under paraphrase and back-translation while reducing accuracy on clean text \citep{kikteva2026show}. Section~\ref{sec:other} gives the comparison. This trade-off is more informative than a ranking on unmodified text alone. A design that varies the rewriting operator, measure family, and corpus together would show which aspects of a representation survive which changes (Section~\ref{sec:open}).

Removing identifying cues and reproducing another writer's cues are different tasks. The mixed impersonation results above show why success at obfuscation cannot be treated as success at imitation. Paraphrase inversion makes this distinction useful for representation learning by treating the information needed to reconstruct an original from its paraphrase as a style-learning signal (Section~\ref{sec:learned}). This suggests studying the interventions by examining what each rewriting changes and whether those changes account for the loss, retention, or transfer of identifying evidence.

\paragraph{When the author is a model} In machine-text detection, style representations trained purely on human authorship transfer to an author population that did not exist at training time. Few-shot detection of machine-generated text with human-authorship style representations works on Llama-2, ChatGPT, and GPT-4 output without training on any generator \citep{riverasoto2024fewshot}. Models have a consistent stylometric footprint that editors lack \citep{shan2026writers}. Lexical, syntactic, and structural features attribute text to a model family and vary within families \citep{kumarage2023neural}. Interpretable stylometric features at ten-sentence length reach a Matthews correlation coefficient (MCC) of 0.870 on a seven-class human/LLM task and locate the difference in greater grammatical standardization of LLM text \citep{przystalski2026stylometry}. Code stylometry separates GPT-4.1 from GPT-4o at 97.56\%, with commenting patterns the strongest cue and signal persisting after comments are stripped \citep{bisztray2025iknow}.

In attribution of machine text to its generating model, the generalization problem recurs. Multilingual machine-text attribution exceeds F1 0.900 in-language and collapses across dissimilar language families, with Russian training transferring better than English \citep{lacava2025authorship}. On 928 books in six languages under domain, author, and language shift no method is best across settings and all degrade under every shift \citep{greco2026multighostbench}. Token rank-transition fingerprints under a small proxy model generalize to unseen generators better than fine-tuned classifiers, a model-idiolect analogue of function-word frequencies \citep{shetty2026who}. Mixed authorship needs segment-local representations. Intermediate versions of a document under progressive AI editing are often harder to detect than either endpoint \citep{bsharat2026opai}. Token-level localization of model contributions beats sentence- and paragraph-level segmentation on the CoAuthor data because authorship changes mid-sentence \citep{lu2026detecting}. Attribution of human--AI collaborative writing is its own emerging task \citep{richburg2024automatic}.

\subsection{What survives shift}\label{sec:survives}

The practical choice is not the feature family with the highest score anywhere, but a method supported under the kind of change expected between reference and questioned texts. Several approaches have supplied that evidence, including punctuation and affix character n-grams, which were the top cross-domain categories \citep{sapkota2015not}. Classical n-gram methods won all four domain-shift settings in a standardized benchmark while losing in distribution \citep{tyo2022state}, while grammar models with masking ranked first on 7 of 12 corpora \citep{nini2026grammar}. In machine-text detection, aggregation over multiple documents recovers detection performance lost to single-document rewriting \citep{riverasoto2025attacks}. The composition of the training data affected transfer in each of the three settings of Section~\ref{sec:diversity}. Several assumptions are not supported by this evidence. Accuracies from different studies should not be compared, nor should a measure family be described as topic-independent. A benchmark's split labels do not necessarily describe what the split varies. Robustness to one shift does not establish robustness to genre or register shift \citep{bagnall2015author, petropoulos2023contrastive}.

\begingroup
\tableformat
\begin{longtable}{L{0.15\linewidth}L{0.29\linewidth}L{0.1\linewidth}L{0.22\linewidth}L{0.15\linewidth}}
\caption{Within-study contrasts at the measure, representation, and protocol layers. Masking rows concern input transformations and, for a learned mask, representation learning; they are not new linguistic measures. Rows that change more than one thing, or report the same system under different metrics, are labeled as such. Contrasts across domains and evaluation conditions are in Table~\ref{tab:factors}; comparisons of scoring and decision rules, including a cross-task successor comparison, are in Table~\ref{tab:readouts}. No cross-study ranking is intended.}\label{tab:generalization}\\
\toprule
\textbf{Layer} & \textbf{Contrast (held constant)} & \textbf{Metric} & \textbf{Baseline $\to$ changed} & \textbf{Source} \\
\midrule
\endfirsthead
\caption[]{(continued)}\\
\toprule
\textbf{Layer} & \textbf{Contrast (held constant)} & \textbf{Metric} & \textbf{Baseline $\to$ changed} & \textbf{Source} \\
\midrule
\endhead
Measure & Distortion off vs.\ on, within topic / cross-topic / cross-genre & accuracy & 80.6$\to$78.2 / 66.2$\to$75.1 / 39.4$\to$58.9 & \citet{stamatatos2017authorship} \\
Measure & Proper-noun vs.\ common-noun masking, PPM, Guardian10 cross-topic & accuracy & 70.41$\to$80.30 / 70.41$\to$69.10 & \citet{sundararajan2018represents} \\
Measure & All untyped vs.\ affix+punct char trigrams, Guardian1 & accuracy & 55.6$\to$57.0, $\sim$1/3 fewer features & \citet{sapkota2015not} \\
Measure & Min.\ frequency 5 vs.\ 200, MNB affix+punct, cross-topic & accuracy & 64.3$\to$73.2 (341 features) & \citet{markov2017improving} \\
Measure & Distinct-symbol replacement of 150 highly frequent words after named-entity, digit, and punctuation preprocessing, MNB & accuracy & 64.3$\to$62.2 & \citet{markov2017improving} \\
Measure & POS masking off vs.\ on, contrastive retrieval, cross-genre / per-genre & Success@8 & 0.225$\to$0.199 / 0.570$\to$0.459 & \citet{fincke2024separating} \\
Measure & No mask vs.\ stopword / POS / learned mask, retrieval & R@8 ($\times100$) & 22.60$\to$10.34 / 19.14 / 24.66 & \citet{kim2025leveraging} \\
Measure & POSNoise off vs.\ on, grammar model, IMDb & accuracy & 0.751$\to$0.831 & \citet{nini2026grammar} \\
Measure & Full text vs.\ content-masked, Siamese RoBERTa, Amazon & F1 & 0.831$\to$0.786 & \citet{zhu2021idiosyncratic} \\
Representation & Separately tuned frequencies vs.\ adjacency network, adaptive word inventory, 10 authors & error & 10.8$\to$8.3 & \citet{segarra2015authorship} \\
Representation & Token-count vs.\ semantic-neighbor normalization, Cosine Delta & F1 & 0.908$\to$0.959 & \citet{eder2022boosting} \\
Representation & Simple fine-tune vs.\ content-negative contrastive, Blog & accuracy & 0.3928$\to$0.4469 & \citet{huertastato2024isolating} \\
Representation & 100K vs.\ 1M vs.\ 5M training authors, paraphrase sensitivity & MRR drop & 0.169$\to$0.237$\to$0.261 & \citet{wang2023authorship} \\
Representation & Content predicts label at train; evaluate matched vs.\ unmatched & accuracy & 94.9$\to$55.3 (chance 50) & \citet{liu2026style} \\
Metric & Same system reported as AUC / accuracy / macro-accuracy (no intervention) & --- & 0.92 / 82.33 / 67.21 & \citet{tyo2022state} \\
Protocol & Fixed truncation vs.\ chunk-and-pool, 5 style models, multilingual verification & rank & MSR first $\to$ StyleDistance first, CISR second & \citet{riverasoto2026steb} \\
Protocol & Random vs.\ topic-controlled split; content-only / char n-gram model & overall & 0.874$\to$0.852 / 0.927$\to$0.926 & \citet{sawatphol2024addressing} \\
Protocol & Length and size unstandardized vs.\ standardized & ranking & model ranking changes; genre ordering stable & \citet{kiefer2026when} \\
Protocol & Machine-text detection, single vs.\ $\ge$5 documents per author under style attack & AUROC & collapse $\to$ re-separation & \citet{riverasoto2025attacks} \\
\bottomrule
\end{longtable}
\endgroup
\section{Open problems}\label{sec:open}

Several of the disagreements in the preceding sections can be stated as experiments left open by the reviewed literature, each with two explanations it would separate. The first concerns rewriting. Section~\ref{sec:llm-shift} reports drops for whole systems under paraphrase, correction, and back-translation, and one channel ablation \citep{khan2026iamnoone}, but none of the reviewed studies varies the rewriting operator, the measure family, and the corpus together. Such a design would separate two explanations of the results in Section~\ref{sec:factors-mediation}: that attribution falls because particular measures (sentence length, punctuation, function-word frequency) are altered, or that it falls because the rewriting changes many measures at once and the classifier depended on their combination. The operators and the content control already exist in the native-language design of \citet{uluslu2026impact}. The missing pieces are the feature decomposition and the authorship task. A related design would run the same set of rewriting and correction operators on one corpus and read the effect on each measure family, which would also settle how much of the loss under light correction in native-language identification carries over to individual authorship.

The second concerns content masking. The gains for shallow models under topic shift and the losses for learned representations under fixed masks (Section~\ref{sec:masking}) admit two readings: that the model class determines whether masking helps, or that what the mask removes does. Separating them needs one corpus with an explicit topic shift, two or more mask types (resource-based and learned), and both a frequency model and a contrastive encoder, so that model class and mask type vary independently. The reviewed studies do not make that factorial comparison. The same design would show whether the cost of masking for learned representations is a cost of the mask or of the resource it depends on, which \citet{kim2025leveraging} raise and do not resolve.

The third concerns what the standard corpora measure. The reviewed studies do not provide a common author-from-topic-only or author-from-metadata-only baseline across IMDb62, the Blog Authorship Corpus, CCAT, Guardian, and the PAN fan-fiction sets, although they continue to be reused, including in 2024 benchmarks for large language models \citep{wen2024aidbench}. On Reuters C50, attribution was unchanged after heavy rewriting \citep{malik2026assistant}, which is a reason to run the baseline there. Whether other corpora behave the same way is what the baseline would show. Two further conditions need controlled benchmarks. Time-shift numbers come from two studies designed for the purpose \citep{cafiero2025this, kiefer2026when} and one applicability study \citep{halvani2019assessing}, and a shared benchmark with controlled gaps would test the prediction of the time-indexed idiolect \citep{nini2023theory} directly. A 37-million-word French resource exists for it \citep{seminck2021corpus}. Formality is routinely named as a register variable and has not been isolated experimentally in the authorship studies reviewed here. The available control found similar measured formality across modes rather than isolating its effect \citep{wang2021mode}. A large share of the 2023--2026 cross-genre results are measured on the HIATUS corpora, whose design we could trace only to a program data sheet \citep{iarpa2024hiatus}. A primary publication of their design would let readers check what those numbers mean.

The fourth concerns the representation layer. Compression models have been tested in PAN evaluations \citep{halvani2018crossdomain} and cross-topic comparisons \citep{sundararajan2018represents}, whereas the per-author fine-tuned language models of \citet{huang2024alms} were evaluated within domain. Comparing both families on the same corpus, with matched within- and cross-domain conditions, would help identify which aspects of their modeling transfer. In the reviewed literature, none of punctuation, readability, idiosyncratic errors, or syntactic n-grams has a dedicated cross-genre or cross-register study: punctuation's reputation as a cross-topic measure rests on typed n-gram categories \citep{sapkota2015not} and has not been isolated from other transcript differences in a speech-register comparison \citep{aggazzotti2025stylometric}, error-based measures have no cross-domain evaluation, and no syntactic n-gram study uses an explicit cross-topic split.

Behind these sits the question of Section~\ref{sec:terms}, whether the signal that stylometry recovers is idiolect or individual register variation \citep{grieve2023register, nini2023theory}. Three existing designs each hold one condition fixed while the author varies: the within-author genre manipulation of the HIATUS corpora, the same-conversation negatives of \citet{wegmann2022same}, and the content-controlled parallel translations of \citet{liu2026style}. Running one representation across all three would show how its performance changes as register, topic, and content are controlled in turn, and identify which author differences persist within and across these conditions. Relating those patterns to either account requires specifying which variation the account attributes to the individual and which to the communicative situation. Two recent findings offer hypotheses for that experiment. When 18 writers summarized the same source text twice, authorial variation arose from ``the combination of a large number of core grammatical constructions'' rather than from isolated unique n-grams \citep{barlow2025corpus}, which favors distributed over rare-feature representations. Per-aspect style directions in activation space share a common backbone and conflict in their residuals \citep{tran2026can}, which would explain uneven transfer across registers if it holds for author identity. A meta-analysis of the present literature is not feasible, because corpora, candidate counts, text lengths, metrics, and shift definitions differ from study to study, and the protocol reversals of Section~\ref{sec:protocol-effects} show that the differences are not small.
\section{Conclusion}\label{sec:conclusion}

The question posed at the outset was whether an identification model recognizes the author or the circumstances in which the author wrote. A benchmark score does not settle it: a model built from content words alone reached AUC 0.95 on topic-controlled subsamples of a benchmark labeled cross-topic, and the representation that leads STEB's retrieval task scores 0.106 on its content-independence probe. The score becomes informative only when the split controls what it claims to vary, and the literature contains three designs that do so: the same authors writing in every genre \citep{iarpa2024hiatus}, different authors writing within the same conversation \citep{wegmann2022same}, and parallel texts whose content is fixed by construction \citep{liu2026style}. Each removes one alternative explanation of a correct attribution. The next step is to run one representation through all three.

This is why we argue that topic-independence is not a property of a stylistic measure. It is a property of a measure together with its encoding, its scoring rule, and the split on which it is evaluated. Function words, character n-grams, and common words all carry measurable topic effects; the same word list normalized differently moved Cosine Delta from F1 0.908 to 0.959; and, in the one study that isolated it, the same backbone, data, and loss trained with two scoring mechanisms gave R@20 of 0.121 and 0.497 \citep{kulumba2026where}. Two decades of asking which features are topic-independent therefore produced conflicting answers by construction, and a ranking of feature families cannot replace evidence about the representation under the domain change it will face.

The consequence for practice is to choose measures and interventions for the shift they must withstand, and to read any intervention by what it leaves in place. Punctuation and affix categories performed best across domains in the character n-gram comparison, and emoji features held their accuracy across subreddits while a classical feature set fell. Content masking helped frequency models under an explicit topic shift but hurt learned representations under fixed masks. The studies differ in enough ways that the reason is open. Topically diverse reference collections and content-controlled training pairs reduced reliance on topic without removing every content-bearing word. On Reuters C50, where each journalist writes a fixed beat, extensive rewriting left attribution accuracy unchanged, which the study's authors attribute to topic remaining tied to author identity; unchanged accuracy alone therefore does not establish that the stylistic signal survived.

Language-model rewriting sharpens the same question. Off-the-shelf paraphrasers defeated a transformer verifier in 83\% and 92\% of cases and targeted impersonation succeeded in 78\% on fan fiction, whereas GPT-4o impersonations failed against AdHominem and LUAR and ChatGPT imitations of 300 reviewers were evidentially null under a likelihood-ratio framework. Removing an author's cues and reproducing them are different tasks, and whether either succeeds depends on how the rewriting was optimized and how the result was evaluated. Paraphrase inversion turns rewriting into a training signal by learning the information needed to reconstruct an original beyond a frozen semantic representation; what that residual contains, and how it relates to the measures of Section~\ref{sec:measures}, is an open empirical question rather than a definition.

Statistical stylometry began as the alternative to manual comparison of distinctive expressions \citep{holmes1998evolution, rudman1997state}. It now separates authors reliably under stated conditions, from eight controlled students to 100,000 blog authors. The within-study contrasts and controlled designs reviewed here are the means of stating those conditions, and of finding out which author differences persist when they change.

\bibliographystyle{plainnat}
\bibliography{references}

\end{document}